\documentclass[journal, twocolumn]{IEEEtran}
\ifCLASSINFOpdf
\else
\fi
\usepackage{times}
\usepackage{epsfig}
\usepackage{graphicx}
\usepackage{amsmath}
\usepackage{amssymb}
\usepackage{enumerate}
\usepackage{subfig}
\usepackage{wrapfig}
\usepackage{amsmath}
\usepackage{caption}
\usepackage{url}
\usepackage{multirow}
\usepackage{multicol}
\begin{document}
%
\title{Deep generative-contrastive networks for facial expression recognition}
%
%
%
\author{Youngsung Kim$^{\dagger}$, ByungIn Yoo$^{\dagger, \ddagger}$, Youngjun Kwak$^{\dagger}$, Changkyu Choi$^{\dagger}$,and Junmo Kim$^{\ddagger}$ %
\thanks{$^{\dagger}$Samsung Advanced Institute of Technology (SAIT) and $^{\ddagger}$Korea Advanced Institute of Science and Technology (KAIST).~%
Corresponding author's email: {\tt\small yo.s.ung.kim@samsung.com}.

}
}
\maketitle

\begin{abstract}
As the expressive depth of an emotional face differs with individuals or expressions, recognizing an expression using a single facial image at a moment is difficult. A relative expression of a query face compared to a reference face might alleviate this difficulty. In this paper, we propose to utilize contrastive representation that embeds a distinctive expressive factor for a discriminative purpose. The contrastive representation is calculated at the embedding layer of deep networks by comparing a given (query) image with the reference image. We attempt to utilize a generative reference image that is estimated based on the given image. Consequently, we deploy deep neural networks that embed a combination of a generative model, a contrastive model, and a discriminative model with an end-to-end training manner. In our proposed networks, we attempt to disentangle a facial expressive factor in two steps including learning of a generator network and a contrastive encoder network. We conducted extensive experiments on publicly available face expression databases (CK+, MMI, Oulu-CASIA, and in-the-wild databases) that have been widely adopted in the recent literatures. The proposed method outperforms the known state-of-the art methods in terms of the recognition accuracy.
\end{abstract}


%
\IEEEpeerreviewmaketitle

\section{Introduction}
Facial expressions are a primary modality to understand the emotional status of an individual. The expression provides a useful contextual clue for social communication~\cite{haxby2000distributed}. However, individuals do not always clearly reveal their facial expressions. When an individual reveals an ambiguous facial expression, a human may have an experience to compare their expression with other expressions observed in past in order to extract their facial expression differences. The related evidence is found in the literature of brain sciences. According to ~\cite{haxby2000distributed, bruce1986understanding, calder2005understanding}, an individual can discern various facial expressions by recalling the memorized face shapes of a shown person. The neural pathways for detecting changeable aspects of faces (e.g., eye movements and emotional expressions)  and for memorizing the unique face shape are separately distributed~\cite{haxby2000distributed, bruce1986understanding}. These  two processes are interacted in the core system of the brain~\cite{haxby2000distributed, calder2005understanding}.

\begin{figure}[t]
\begin{center}
\hspace*{-1.5em} 
\vspace*{-1.5em} 
\includegraphics[width=0.98\linewidth]{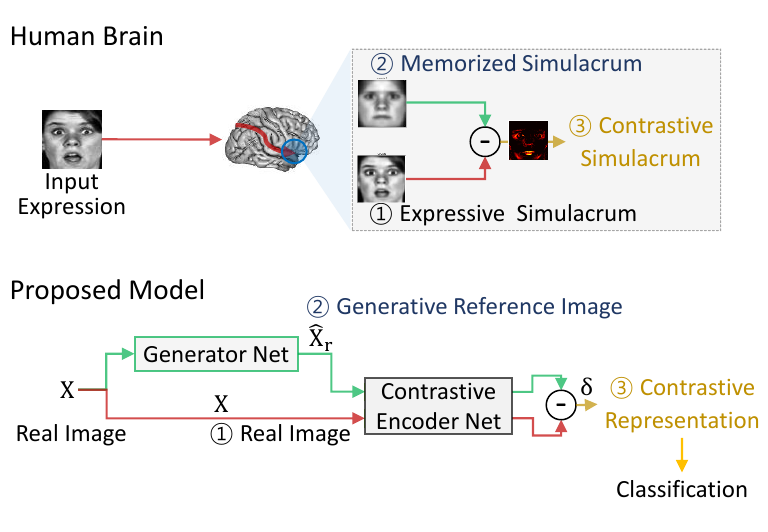}
\end{center}
   \caption{Overview of our proposed architecture. A similar procedure with the proposed architecture might be observed in a human brain. A given expressive face is compared with the memorized facial shape in the brain. In the proposed networks, a contrastive representation $\boldsymbol{\delta}$ of the given image $\mathbf{X}$ is compared with that of a reference image $\mathbf{\hat{X}}_r$ which is estimated by a generator network.}
\label{fig:overview_concept}
\end{figure}

We attempt to utilize a reference face image that indicates the memorized unique face in the brain to discriminate a facial expression input in a deep neural network framework (see Figure~\ref{fig:overview_concept}). We assume that a distinctive expression feature can be extracted from the \textit{contrastive characteristics} between a given image and the reference image. The reference image for an individual identity, however, is not always available in the wild. We start from the assumption that a generative model can infer the reference image from the given image. If a single image is given, the reference image is generated using the generative (encoder-decoder) networks: a generative reference image. The next required process is to model the \textit{contrastive characteristics} mentioned above using deep neural networks.

One of main concerns is to find out how to extract or encode the contrastive representation.
In our proposed network, two representation learning models are included: 1) disentangling of expressions and 2) explaining of a distinctive expression feature.
In general, deep networks disentangle multiple variation factors of an input image~\cite{Bengio:2013:RLR}.
Several unknown or unintended factors can be revealed in the networks and a useful factor is selected by a proper objective. In this paper, we attempt to disentangle directly the intended attribute: a facial expression. Hence, disentangling of expression is conducted in two steps. First, through learning a generator network that estimates the reference image, some latent representations related to an expression can be eliminated. This estimated reference image is used to measure the distinctive expression feature  in a latent space. In a later part, disentangling is assisted by contrastive metric learning and a reconstruction learning.

From the approach in the literature~\cite{liu2014learning}, gradual changes of facial expressions are utilized to extract temporal information along the multiple frames. This multiple images (video) based model has abundant information of the expression transition, which can be used for the recognition. In this paper, we focus on exploring the representation from a pair of the generative image and the given image. Our proposed framework could be easily extended to utilize multiple frames as well.

In this paper, we attempt to answer to a few questions quantitatively and qualitatively: 1) Is the generative reference image useful for the discriminative task? 2) How generative networks are controlled by contrastive metric learning for a discriminative purpose? and 3) How does facial generation affect expression recognition?

The main contributions of this paper are as follows:
\begin{itemize}
\item We combine encoder-decoder networks and convolutional neural networks into a unified network that simultaneously learns to generate, compare, and classify an input data.
\item We show that the contrastive representation trained with contrastive metric learning and reconstruction learning is useful to achieve a better discriminative performance for a facial expression recognition task.
\item We show that the proposed (single image based) method outperforms the state-of-the-art methods including the multiple images based approach in terms of facial expression recognition accuracy.
\end{itemize}

\begin{figure*}[hbt]
\begin{center}
\hspace*{-1.7em} 
\subfloat[CNN]{\includegraphics[width=0.17\linewidth]{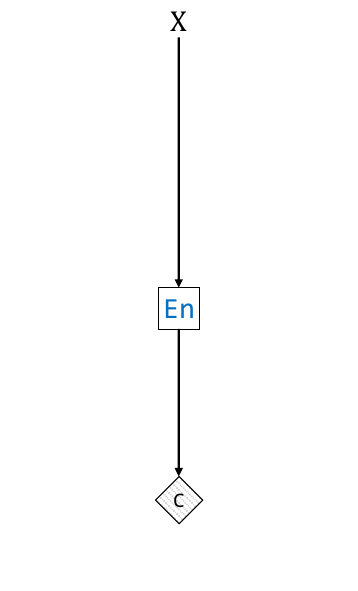} }
\hspace*{-2.5em} 
\subfloat[metric-CNN]{\includegraphics[width=0.17\linewidth]{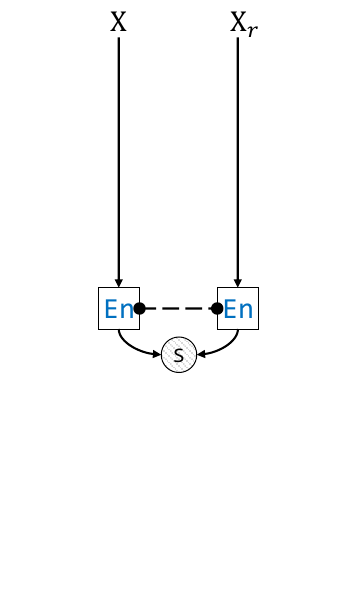} }%
\hspace*{-1.9em} 
\subfloat[$\textrm{GCNet}_{S0R0}$]{\includegraphics[width=0.17\linewidth]{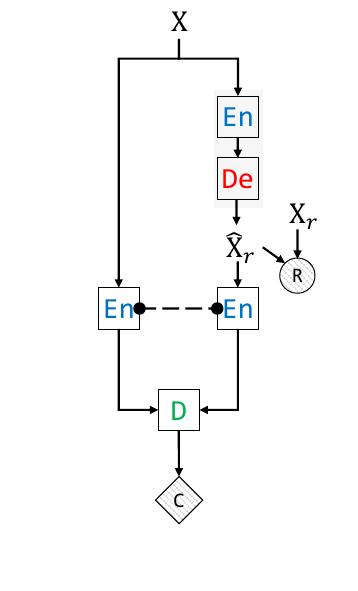} }
\hspace*{-1.1em} 
\subfloat[$\textrm{GCNet}_{S1R0}$]{\includegraphics[width=0.17\linewidth]{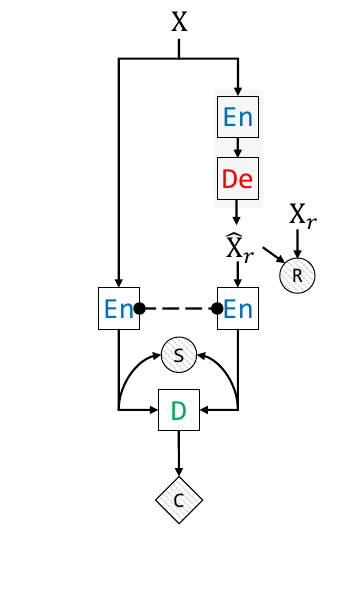} }%
\hspace*{-1.3em} 
\subfloat[$\textrm{GCNet}_{S0R1}$]{\includegraphics[width=0.17\linewidth]{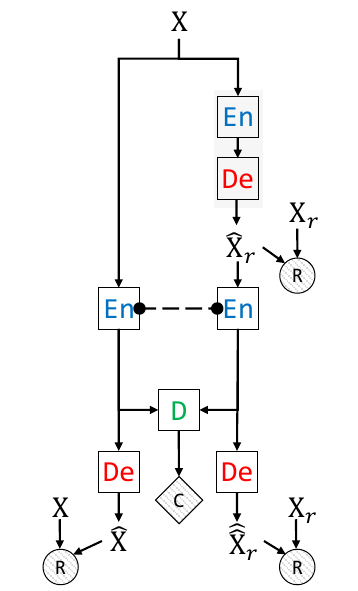} }
\hspace*{-1.1em} 
\subfloat[$\textrm{GCNet}_{S1R1}$]{\includegraphics[width=0.17\linewidth]{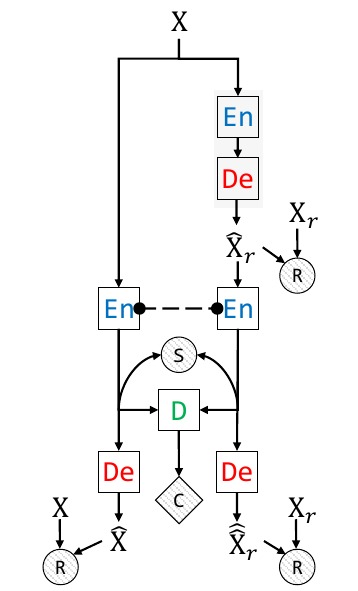} }%
\hspace*{-0.3em} 
\subfloat{\includegraphics[width=0.15\linewidth]{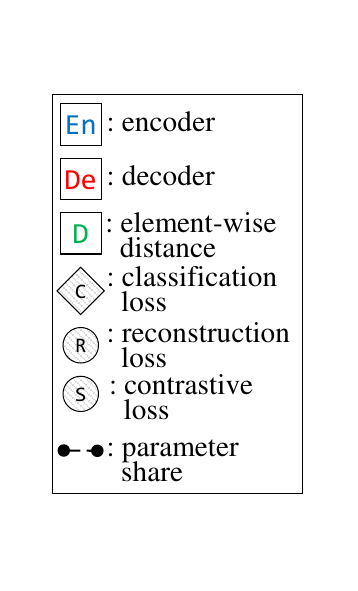} } %
\end{center}
\vspace*{-0.5em}
   \caption{Architecture overviews of our proposed networks derived from (a) and (b) in a training phase:
   (a) CNN,
   (b) metric-learning-CNN with a paired input $\{\mathbf{X}, \mathbf{X}_r\}$ where $\mathbf{X}$ is a given expression and $\mathbf{X}_r$ is a reference expression, 
   (c) $\textrm{GCNet}_{S0R0}$: contrastive representation using a generative image ($\hat{\mathbf{X}}_r$) is adopted for a discriminative task (where $\hat{\mathbf{X}}_r$ is generated via encoder-decoder networks),
   (d) $\textrm{GCNet}_{S1R0}$: a contrastive metric loss ($_S$) is added on $\textrm{GCNet}_{S0R0}$,
   (e) $\textrm{GCNet}_{S0R1}$: decoder networks with a reconstruction loss ($_R$) are added on $\textrm{GCNet}_{S0R0}$ for a better representation ($\hat{\mathbf{X}}$ is a reconstructed sample of the given expression and $\hat{\hat{\mathbf{X}}}_r$ is a reconstructed sample of the generated reference image),
   and (f)  $\textrm{GCNet}_{S1R1}$: a contrastive metric loss ($_S$) and a reconstruction loss ($_R$) are added.}
\label{fig:overview}
\end{figure*}

\section{Related works}
Facial expression recognition has been studied over decades. Several different approaches exist that are based on local feature extraction, facial action units (FAUs), temporal information, and convolutional neural networks. The local feature-based methods such as the Gabor filter, LBP, HOG, and BoW are the most common and widely studied to extract good visual features \cite{bartlett2005recognizing, Klaser_aspatio-temporal2008, Zhao:2007:DTR}. In the FAU based methods \cite{liu2013aware, liu2015inspired}, FAUs are detected and analyzed to classify an expression. This is mainly based on the facial action coding system (FACS) proposed by Paul Ekman \cite{ekman:friesen:1978}. Temporal information-based methods \cite{liu2014learning, kacem2017novel}  utilize multiple images. These methods, however, achieve \textit{limited} recognition accuracy performance because the designed features lost some information. To overcome the insufficient representations of the hand-crafted features, deep learning based methods have been recently adopted. An ensemble of two deep networks models that handle temporal information including appearance and geometric features has been proposed~\cite{jung2015joint}. A simple convolutional neural network has been used to analyze the FAU in the learned filter of the networks~\cite{khorrami2015deep}. To obtain discriminative spatiotemporal representation, facial action parts detection is performed using 3D-CNN~\cite{liu2014deeply}. However, it shows limited performance when compared to the state-of-the-art methods. This is because those CNN-based methods still could not show a good enough representation of a facial expression.

Another deep learning framework has been proposed to take advantage of the discriminative and generative models for realizing a better generalization performance. Traditionally in generative networks such as the autoencoder, a popular approach is that the entire stack of encoders is finetuned using pre-trained autoencoders in a layer-wise manner for discriminative purposes. Recently, a generative model was simultaneously learned with a discriminative model. In generative adversarial networks (GANs)~\cite{Goodfellow:NIPS2014}, the generative model is learned against an adversary and a discriminative model that learns to determine whether a sample is from the model distribution or data distribution. The stacked what-where auto-encoders (SWWAEs)~\cite{zhaoMGL15} integrate discriminative and generative learning pathways and provide a unified approach to supervised, semi-supervised, and unsupervised learning. In this paper, we deploy a generative model with discriminative learning as well. We are mainly focusing on investigating a contrastive representation of a facial expression that is optimized with appropriate objectives.


\section{Generative and contrastive facial representation learning}
Consider an input image matrix $\mathbf{X}\in \mathbb{R}^{h \times w}$ and a reference image matrix $\mathbf{X}_r\in \mathbb{R}^{h \times w}$ that are elements of an image set $\mathcal{I}$ and the data space $\mathcal{X}$.  The corresponding expression labels denoted by $\{y, y_r\} \in \mathbb{R}$ are elements of a label set $\mathcal{Y}$. In the real world, an expressive face might be changed from a ground face (due to emotional changes that incur facial muscle movements \cite{wang2013capturing}). We define a relationship between two images with a hidden  factor denoted by $\boldsymbol{\epsilon} \in \mathbb{R}^{h \times w}$ formally as follows:
\begin{equation}\label{eq:additive_epsilon}
  \mathbf{X} := \mathbf{X}_{r} + \boldsymbol{\epsilon}
\end{equation}
where the addition indicates operations for facial expression change\footnote{Since the change of expression should be measured in the same subject, we assumed that a hidden expression factor is represented within the same subject, i.e. if a subject term $s$ is added at the Equation~(\ref{eq:additive_epsilon}): $\mathbf{X}_{s} := {\mathbf{X}_{s}}_{r} + \boldsymbol{\epsilon}$. In this paper, we omit the term $s$ for a simplicity in the notation.}.

\subsection{Contrastive representation}

\begin{figure*}[hbt]
\begin{center}
  \includegraphics[width=0.905\linewidth]{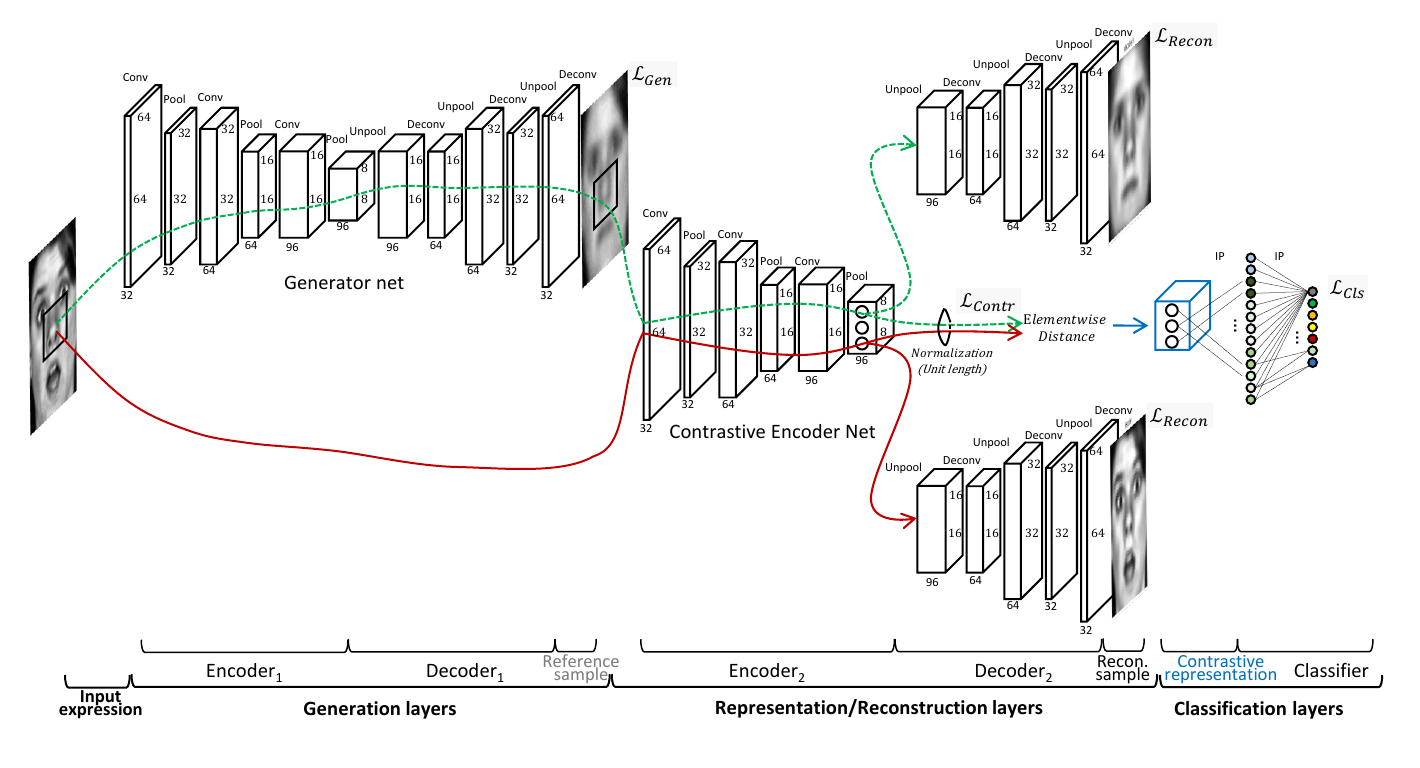}
  \vspace*{-1.5em}
\end{center}
  \vspace*{-0.5em}
   \caption{The overall architecture of the proposed networks (Figure~\ref{fig:overview}~(f) in detail).
    Two-way data flows starting from a given expression image exist over the generation layers and the representation/reconstruction layers.
    A dashed-line arrow (green color) depicts a flow to represent processing using a reference image generated by the generator networks.
    A solid-line arrow (red color) depicts a flow using the given expression image. Loss functions $\mathcal{L}_{Cls}$, $\mathcal{L}_{Gen}$, $\mathcal{L}_{Contr}$, and $\mathcal{L}_{Recon}$ are required for learning. In a test phase, $\mathsf{Decoder}_2$ layers are not required. }
\label{fig:overview_detail}
\end{figure*}

As a facial expression is not always apparently represented as an absolute value, aspects of expression change obtained by comparing with the reference image might be useful. An expression image with a very small change could be recognized via difference maps (e.g., a pixel-wise distance and optical flows). A simple approach is to compare image pixels of the faces. However, owing to distortions between the images (e.g., distortions by an affine transform), comparing the images at the pixel level is not effective. For example, a small translation in the image level might return large pixel-wise errors even though a human face shows no expression changes. The representation of the difference can be better extracted at the feature level that offer an invariance towards distortions than at the pixel level.

We employ a contrastive representation to extract a distinctive feature from a pair of expression images in a latent space. We refer to the data space $\mathcal{X} := \{\mathbf{X} \in \mathbb{R}^{h \times w}~|~\mathsf{En}(\mathbf{X};\bullet) \in \mathcal{Z} \}$ where $\mathcal{Z}$ is the latent space and $\mathsf{En}(\mathbf{X};\bullet)$ (an encoder) denotes a transform function used to map the data space $\mathcal{X}$ to the latent space $\mathcal{Z}$: $\mathbb{R}^{h \times w} \rightarrow \mathbb{R}^{p}$ with the learnable parameters $\bullet$. In the latent space, the contrastive representation $\boldsymbol{\delta}$ can be defined as follows:
\begin{equation}\label{eq:contrastive_representation}
  \boldsymbol{\delta} := d(\mathsf{En}(\mathbf{X};\theta_e),  \mathsf{En}(\mathbf{X}_{r};\theta_e)),
  \vspace*{-0.5em}
\end{equation}
where $d(\cdot, \cdot) \in \mathbb{R}^p$ denotes an element-wise distance, and $\theta_e$ denotes the learnable parameters. In this paper, we adopt a normalized distance $\boldsymbol{\delta}_j = {\|
\frac{\mathsf{En}(\mathbf{X};\theta_e)_j }
{\| \mathsf{En}(\mathbf{X};\theta_e) \|} - \frac{ \mathsf{En}(\mathbf{X}_{r};\theta_e)_j}{\|  \mathsf{En}(\mathbf{X}_{r};\theta_e)  \|}  \|} \in \mathbb{R}~\forall j$ for the $j$-th element of the representation vector $\boldsymbol{\delta}\in \mathbb{R}^p$ where $j=1,2,\ldots,p.$ and $\|\cdot\|$ denotes the $L^2$-norm.

The contrastive representation $\boldsymbol{\delta} \in \mathbb{R}^p$ is an input for a discriminative task (please refer to ``\textsf{\footnotesize{D}}'' in Figure~\ref{fig:overview} (c), (d), (e), and (f)). This representation is extracted from \textsf{\footnotesize{Contrastive Encoder Net}} and adopted as the input at \textsf{\footnotesize{Classification layers}} as shown in Figure~\ref{fig:overview_detail}.

\subsection{Generative reference image}
A pair of images, the query face and the reference face  $\{\mathbf{X}, \mathbf{X}_{r}\}$, may not be available at the same time in the test phase.
In this paper, therefore, we propose to generate the reference face using the generative networks.
As a human keeps a less-expressive (or neutral-like) face most of time, that face image could be considered as the reference image.
We adopt the convolutional encoder-decoder networks in this paper, to estimate the reference face transformed from an expressive face.
This is because the concept of the denoising auto-encoder (DAE)~\cite{Bengio:2013} matches with this situation\footnote{In the DAE, a term corresponding to corruption, i.e., a Gaussian distribution, added to the original input is eliminated via learning. An observed random variable $X$ is corrupted into $\tilde{X}$ using the known conditional distribution $C(\tilde{X}~|~X)$ in order to train the autoencoder to estimate the reverse conditional $P(X~|~\tilde{X})$. In this paper, we assume that an observed random variable, an expressive face, $X$ is corrupted from $X_r$, a reference face, using conditional distribution $C(X~|~X_r)$. The generative networks is used to estimate the reverse conditional $P({X}_r~|~X)$. We assume that the term corresponding to corruption should not be limited to a specific probability distribution.}.
By disentangling facial expressive factors, hence, information that is irrelevant or negligible use for the discriminative purposes could be discarded~\cite{Bengio:2013:RLR}.

Consider $\mathsf{De}(\mathbf{X};\bullet)$ (a decoder) be a transform function used to map the latent space $\mathcal{Z}$ to the data space $\mathcal{X}$: $\mathbb{R}^{p} \rightarrow \mathbb{R}^{h \times w}$ with the learnable parameters $\bullet$. Formally, a generative reference image ${\mathbf{\hat{X}}_r}$ of the input image $\mathbf{X}$  is estimated using a generator network $\mathsf{G}$ which consists of the encoder $\mathsf{En}$ and the decoder $\mathsf{De}$ with the learnable parameters as follows,
\begin{equation}\label{eq:generator}
     {{\mathbf{\hat {X}}}_r}  :=  \mathsf{G}({\mathbf{X}};\psi) :=  \mathsf{De}(\mathsf{En}({\mathbf{X}; \psi_e});\psi_d),
\end{equation}
where the learnable parameters $\psi$ consists of the learnable parameters $\psi_e$ and $\psi_d$.


\subsection{Networks learning}
The parameters in the networks are learned with multiple loss functions. Formally, the objective loss function to minimize is written as follows:
\begin{align}\label{eq:lossfunction_three}
\mathcal{L}(\phi, \psi, \theta_e, {\theta_d}_i, {\theta_d}_r)  &  =  \mathcal{L}_{Cls}(\phi) + \lambda_{G} \mathcal{L}_{Gen}(\psi) \\ \nonumber
                                                                                                                &  + \lambda_{S} \mathcal{L}_{Contr}(\theta_e) +  \mathcal{L}_{Recon}( {\theta_d}_i, {\theta_d}_r),
\end{align}
\noindent where $\mathcal{L}_{Cls}$ with the learnable parameters $\phi$ denotes a discriminative loss function, $\mathcal{L}_{Gen}$ denotes a generative loss function, $\mathcal{L}_{Contr}$ denotes a contrastive loss function, $\mathcal{L}_{Recon}$ denotes a reconstruction loss function, $\lambda_{G}, \lambda_{S}\in \mathbb{R}$ indicate the weight for the loss functions.

\subsubsection{Loss for discriminative learning}
The main purpose of the proposed networks is to classify a facial expression of the given input. For the discriminative objective $\mathcal{L}_{Cls}$ (please refer to ``\textsf{\footnotesize{C}}'' in Figure~\ref{fig:overview} (c), (d), (e), and (f)), we adopt the cross entropy loss function which is widely used for the classification task.

\subsubsection{Loss for generative learning}
The main purpose of a generative loss is to generate the reference image. The generative loss $\mathcal{L}_{Gen}$ can be represented as follows:
\begin{equation}
  \mathcal{L}_{Gen}(\psi)   :=  \frac{1}{2}\|\mathbf{X}_r - \mathsf{G}({\mathbf{X}};\psi)  \|_{2}^2.
\end{equation}

\noindent where the learnable parameters $\psi$ is used to estimate the target reference image $\mathbf{X}_r$ and a pair $\{\mathbf{X}, \mathbf{X}_r\}$ is from the same individual.
In this paper, the aim of the generative image is not at the better visual quality on human eyes, but at a source image of the contrastive encoder networks.
Hence, abundant supervised conditions (e.g. an identity label and a expression label) are not applied to the loss function in generative learning.

For learning the contrastive representation, two objectives are deployed in the proposed networks: the first objective $\mathcal{L}_{Contr}$ is for contrastive metric learning to enlarge or to diminish the distance between the two latent representation vectors, and the second one $\mathcal{L}_{Recon}$ is for reconstruction learning. 

\subsubsection{Loss for \textit{contrastive metric learning}.}
The loss $\mathcal{L}_{Contr}$ aims to optimize a similarity between two encoded representations $\{\textit{encoded real input image}~\mathsf{En}(\mathbf{X};\theta_e)$, $ \textit{encoded generative reference image}~\mathsf{En}(\mathbf{\hat{X}}_{r};\theta_e)=\mathsf{En}(\mathsf{G}(\mathbf{X};\psi);\theta_e)\}$  according to their expression labels (please refer to ``\textsf{\footnotesize{S}}'' in Figure~\ref{fig:overview} (d) and (f)).  If the expression labels of ${\mathbf{X}}$ and ${\mathbf{\hat {X}}}_r$ are not identical, the function optimizes to obtain dissimilar features within a predefined margin; if the expressions are identical, it optimizes to similar features. Hence, the contrastive loss \cite{hadsell2006dimensionality} $\mathcal{L}_{Contr}$ is adopted in the latent space as follows:
\begin{equation}
\begin{split}
  &\mathcal{L}_{Contr}(\theta_e, \psi)\\
                 &:=\alpha  \frac{1}{2}   \{ \max(0, m - S(\mathsf{En}(\mathbf{X};\theta_e), \mathsf{En}(\mathsf{G}(\mathbf{X};\psi);\theta_e)) ) \}^2 \\
                 &+  (1-\alpha)  \frac{1}{2}   \{ S(\mathsf{En}(\mathbf{X};\theta_e), \mathsf{En}(\mathsf{G}(\mathbf{X};\psi);\theta_e)) \}^2,
\end{split}
\end{equation} 
where $\alpha=1$ if the labels $\{{y}, {{{y}}}_r\}$ of a pair $\{\mathbf{X}, \mathsf{G}(\mathbf{X};\psi)\}$ are not the same, $\alpha=0$ otherwise, $S(*, \bullet) :=
\| *-\bullet \|_{2} \in \mathbb{R}$ is a distance measure (low values of $S$ for similar pairs and vice versa), and $m >0$ is a margin.

To find multi-dimensional contrastive representation corresponding to the input image, preserving the data-generating distribution might be useful.
A process of image reconstruction is known ``to capture the structure of the data-generating distribution"~\cite{Bengio:2013:RLR}.

\subsubsection{Loss for reconstruction learning}
The main purpose of a reconstruction loss is to supplement to the contrastive representation learning (please refer to ``\textsf{\footnotesize{R}}'' in Figure~\ref{fig:overview} (e) and (f)). The reconstruction loss $\mathcal{L}_{Recon}$ can be represented as a weighted summation of two reconstruction terms as follows:
\begin{equation}\label{eq:lossfunction}
 \mathcal{L}_{Recon}( {\theta_d}_i, {\theta_d}_r) =  \lambda_{R,i} \mathcal{L}_{Recon, i}({\theta_d}_i)
+  \lambda_{R,r} \mathcal{L}_{Recon, r}({\theta_d}_r),
\end{equation}

\noindent where of the subscript $_{\bullet} \in \{i,r\}$, indicates a target: $i$ for the \textit{i}nput image and $r$ for the \textit{r}eference image respectively. The decoders with the parameters ${\theta_d}_i$ and ${\theta_d}_r$ to find the facial expression-generating representation in the contrastive encoder layers learns with the losses $\mathcal{L}_{Recon,i}$ and $\mathcal{L}_{Recon,r}$ respectively as follows,
\begin{equation}
  \mathcal{L}_{Recon,i}(\theta_e,{\theta_d}_i )   := \frac{1}{2}\|\mathbf{X} - \mathsf{De}(\mathsf{En}(\mathbf{X};\theta_e);{\theta_d}_i)  \|_{2}^2,
\end{equation}
\begin{equation}
  \mathcal{L}_{Recon,r}(\psi,\theta_e,{\theta_d}_r)   := \frac{1}{2}\|\mathbf{X}_r - \mathsf{De}(\mathsf{En}(\mathsf{G}(\mathbf{X};\psi);\theta_e)) ;{\theta_d}_r) \|_{2}^2,
\end{equation}
where a target (real) reference image $\mathbf{X}_r$ is given and a generative reference image ${{\mathbf{\hat {X}}}_r} $ is estimated by $\mathsf{G}({\mathbf{X}};\psi) = \mathsf{De}(\mathsf{En}({\mathbf{X}; \psi_e});\psi_d)$.

\section{Experiments}
In this section, we describe the experiments conducted to compare the proposed method with the state-of-the-arts on publicly available face expression databases (CK+, MMI, and Oulu-CASIA) that are widely adopted in the literatures ~\cite{guo2012dynamic, jain2011facial, jung2015joint, khorrami2015deep, Klaser_aspatio-temporal2008,
liu2013aware,liu2015inspired, liu2014deeply, liu2014learning, conf/iccvw/PtuchaTS11, wang2013capturing, zhao2011oulucasia, Zhao:2007:DTR, zhong2012learning}. Additionally, we show an experiment on several in-the-wild databases (RAF~\cite{Li_2017_CVPR}, FER2013~\cite{GOODFELLOW201559}, and SFEW~\cite{Emotiw2015}).

\subsection{Networks model and settings}
All models used in different databases share exactly the same architecture (shown in Figure~\ref{fig:overview_detail}), including encoder-decoder networks depicted in Table~\ref{table:AE_detail}. All parameter settings are shared through the databases with the same value. The encoder-decoder networks in Table~\ref{table:AE_detail} are pre-trained with the reconstruction task using the CASIA-WebFace database~\cite{Yi2014CASIA}, and three convolutional layers in the encoder are adopted at $\textrm{Encoder}_{1}$ ($\mathsf{En}(\bullet;\psi_e)$) of the proposed generative-contrastive networks (GCNet) shown in the Figure \ref{fig:overview_detail}. The baseline CNN consisting of three convolutional layers and two inner-product (FC) layers are pre-trained with the identification task using the same database, and convolutional layers are adopted at $\textrm{Encoder}_{2}$ ($\mathsf{En}(\bullet;\theta_e) $). During the training of the proposed networks, the learning rate at layers of the decoder networks is set to $10$ during fine-tuning. The number of outputs at the first  fully-connected layer (inner-product) is empirically determined by $(0.5*\text{Wsize})^2*\text{nlayers}/(2^{\text{nlayers}})$ where we set $\text{Wsize}=64$, $\text{nlayers}=3$. This is intended that a dimensionality of the vector decreases smoothly as the number of (conv./pool) layers increases. $\frac{1}{2^\text{nlayers}}$ is related to a pooling size ($\frac{1}{2}$) at each layer. The dropout is applied before this fully-connected layer with a ratio of $0.5$. After the FC-layer, a softmax layer is connected with the number of outputs corresponding to the number of classes. We arbitrarily set $\lambda_{S}=1 , \lambda_{G}=1,  \lambda_{R,r}=0.25, \lambda_{R,i}=0.25$ for each loss function. The maximum iteration is set to $ 3\times 10^5$.

Our models are trained with `Nesterov' optimization using an `inverse' learning policy, a base learning rate of $0.001$, a momentum of $0.9$, a gamma term of 0.75, a weight decay of $0.0001$, and a mini-batch size of $64$. The proposed network model is implemented on \textit{Python} and the deep learning framework \textsf{Caffe} and run using the NVIDIA Tesla K80 GPU.

To avoid over-fitting, we applied data augmentation during the training phase. We used input images on a gray level (1 channel) where a facial region is cropped, normalized based on 5 points (eyes, the end of a nose, and two ends of lips) and resized into $66 \times 66$. The resized image is cropped again with the size of $64 \times 64$ at a random location. Each cropped image is manipulated using 2D affine transform such as scaling, rotation, random horizontal flipping, and intensity multiplication, in addition to contrast-limited-adaptive histogram equalization (this is also applied in the test phase).

\subsection{Databases and protocols}
  \subsubsection{CK+ Database \cite{lucey2010CKextended}} This database is widely adopted in the benchmark for facial expression recognition tasks. This database consists of 593 sequences with 123 individuals. The images are captured expression transitions from a neutral face to peak facial expression acted by an individual. The 327 valid sequences with 118 individuals that maintain discrete emotion labels such as ``Anger, Contempt, Disgust, Fear, Happy, Sad, and Surprise" are adopted for an experiment. We divide the valid sequences into ten different subsets with individual-independent way. According to individual ID in the database, individuals are grouped by sampling in ID ascending order with ten even intervals first. One subset out of ten subsets is used for validation (test), the remains are used for training. This procedure is repeated ten times. This subject-independent 10-fold cross-validation follows the previous works \cite{jung2015joint, liu2014learning}.
  \subsubsection{MMI Database \cite{valstar2010induced}} This database consists of 312 sequences from 30 individuals with six basic expressions (Contempt included in the CK+ database is excluded). We selected 205 sequences captured in a front view. Each sequence starts from a neutral face, and shows a peak expression within a single expression type in the middle of the sequence. At the end, it returns to a neutral face again. As a peak expression frame number is not given, we selected it manually. Similar to the CK+ database settings, we divided the MMI database into ten different individual independent subsets. Consequently, 10-fold cross validation was conducted. This database includes individuals who pose expressions non-uniformly, wear glasses/caps, and have mustaches/head movements. Therefore, the facial expression recognition task is relatively challenging. Moreover, the small number of sequences and individuals makes it difficult to achieve a good generalization performance. This database could be suitable to measure the recognition performance in realistic situations when compared to other databases.
  \subsubsection{Oulu-CASIA VIS Database \cite{zhao2011oulucasia}} This database consists of 480 image sequences with 80 individuals. This database is captured under the visible (VIS) normal illumination conditions and is a subset of Oulu-CASIA NIR-VIS database. Each individual poses six basic expressions similar to MMI database. Similar to the CK+ database, the sequence starts from a neutral face and ends with peak facial expression within a same emotion category. As done with the two databases above, individual-independent 10-fold cross-validation is conducted.

\begin{table}[!htb]
\begin{center} {
\begin{tabular}{|c|}
\hline
Encoder (3 convolutional layers)\\
\hline
 ($5\times5$, $32$) Conv. BNorm, ReLU, ($2\times2$) MaxPool \\
 ($3\times3$, $64$) Conv. BNorm, ReLU, ($2\times2$) MaxPool \\
 ($3\times3$, $96$) Conv. BNorm, ReLU, ($2\times2$) MaxPool \\
\hline
Decoder (3 de-convolutional layers)\\
\hline
 ($2\times2$) Max\textit{Un}Pool, ($3\times3$, $64$) \textit{De}Conv. BNorm ReLU \\
 ($2\times2$) Max\textit{Un}Pool, ($3\times3$, $32$) \textit{De}Conv. BNorm, ReLU  \\
 ($2\times2$) Max\textit{Un}Pool, ($5\times5$, $1$) \textit{De}Conv. BNorm, ReLU  \\
\hline
\end{tabular} }
\end{center}
\vspace*{-0.5em}
\caption{Details of the convolutional encoder-decoder layers embedded in the proposed networks. The Max\textit{Un}Pool layer is adopted from the literature~\cite{noh2015learning}. An encoder part consists of three convolutional layers (Conv.) which is followed by Batch Normalization (BNorm), ReLU, and Max Poooling layers. Correspondingly, a decoder part consists of three de-convolutional (transposed convolutional) layers. In a Conv and DeConv. layers, ($5\times5$, $32$) indicates that there is 32 sets of 5$\times$5 filters. In MaxPool and MaxUnPool layers, ($5\times5$) indicates a pooling window size.  }
\label{table:AE_detail}
\end{table}

\subsection{Quantitative results}\label{sec:quantitative}
Among all the compared databases, the proposed methods outperform the state-of-the-art methods including handcraft based methods (LBP-TOP~\cite{Zhao:2007:DTR} and HOG 3D~\cite{Klaser_aspatio-temporal2008}), video-based methods (MSR~\cite{conf/iccvw/PtuchaTS11}, TMS~\cite{jain2011facial}, STM-ExpLet~\cite{liu2014learning}, and DTAGN-Joint~\cite{jung2015joint}) that utilize temporal information, FAU inspired methods (AURF~\cite{liu2013aware}, AUDB~\cite{liu2015inspired}), and CNN-based methods (3D-CNN~\cite{liu2014learning}, 3D-CNN-DAP~\cite{liu2014learning}, zero-bias CNN+AD~\cite{khorrami2015deep}, and DTAGN-Joint~\cite{jung2015joint}).

\begin{table}[!htb]
  \begin{center}\footnotesize{
  \begin{tabular}{|l|c|}
  \hline
  Method & Accuracy ($\%$) \\
  \hline\hline
  LBP-TOP~\cite{Zhao:2007:DTR} & 88.99 \\
  HOG 3D~\cite{Klaser_aspatio-temporal2008} & 91.44 \\
  \hline
  MSR~\cite{conf/iccvw/PtuchaTS11} & 91.4 \\
  TMS (4-fold)~\cite{jain2011facial} & 91.89 \\
  STM-ExpLet~\cite{liu2014learning} & 94.19 \\
  DTAGN-Joint~\cite{jung2015joint} & 97.25 \\
  traj. on S+(2; n)~\cite{kacem2017novel} &  96.87 \\
  \hline
  3D-CNN~\cite{liu2014deeply} & 85.9 \\
  3D-CNN-DAP~\cite{liu2014deeply} & 92.4 \\
  \hline
  CNN (baseline) & 96.94\\
  Ours ($\textrm{GCNet}_{S0R0}$) & \textbf{97.08}\\
  Ours ($\textrm{GCNet}_{S1R0}$) & \textbf{97.83}\\
  Ours ($\textrm{GCNet}_{S0R1}$) & \textbf{97.53}\\
  Ours ($\textrm{GCNet}_{S1R1}$) & \textbf{97.93}\\
  \hline
\end{tabular} }
\end{center}
\vspace*{-1.0em}
\caption{Averaged recognition accuracy ($\%$) on the CK+ database, 7 expressions.}
\label{table:CK+seven}
\end{table}

\begin{table}[!htb]
  \begin{center}\footnotesize{
  \begin{tabular}{|l|c|}
  \hline
  Method & Accuracy ($\%$) \\
  \hline\hline
  AURF~\cite{liu2013aware} & 92.22 \\
  AUDB~\cite{liu2015inspired} & 93.70 \\
  Zero-bias CNN+AD~\cite{khorrami2015deep} & 96.4 \\
  FN2EN~\cite{ding2017facenet2expnet} & 96.8\\
  \hline
  CNN (baseline) & 95.47\\
  Ours ($\textrm{GCNet}_{S0R0}$) & \textbf{95.74}\\
  Ours ($\textrm{GCNet}_{S1R0}$) & \textbf{96.75}\\
  Ours ($\textrm{GCNet}_{S0R1}$) & \textbf{96.50}\\
  Ours ($\textrm{GCNet}_{S1R1}$) & \textbf{97.28}\\
  \hline
\end{tabular} }
\end{center}
\vspace*{-1.0em}
\caption{Averaged recognition accuracy ($\%$) on the CK+ database, 8 expressions.}
\label{table:CK+eight}
\end{table}

\begin{table}[!htb]
  \begin{center} \footnotesize{
  \begin{tabular}{|l|c|}
  \hline
  Method & Accuracy ($\%$) \\
  \hline\hline
  LBP-TOP~\cite{Zhao:2007:DTR} & 59.51 \\
  HOG 3D~\cite{Klaser_aspatio-temporal2008} & 60.89 \\
  \hline
  ITBN~\cite{wang2013capturing} & 59.7 \\
  CSPL~\cite{zhong2012learning} & 73.53 \\
  STM-ExpLet~\cite{liu2014learning} & 75.12 \\
  DTAGN-Joint~\cite{jung2015joint} & 70.24 \\
  traj. on S+(2; n)~\cite{kacem2017novel} & 79.19 \\
  \hline
  3D-CNN~\cite{liu2014deeply} & 53.2 \\
  3D-CNN-DAP~\cite{liu2014deeply} & 63.4 \\
  \hline
  CNN (baseline) & 77.68\\
  Ours ($\textrm{GCNet}_{S0R0}$) & \textbf{76.20}\\
  Ours ($\textrm{GCNet}_{S1R0}$) & \textbf{78.86}\\
  Ours ($\textrm{GCNet}_{S0R1}$) & \textbf{77.00}\\
  Ours ($\textrm{GCNet}_{S1R1}$) & \textbf{81.53}\\
  \hline
\end{tabular} }
\end{center}
\vspace*{-1.0em}
\caption{Averaged recognition accuracy ($\%$) on the MMI database, 6 expressions.}
\label{table:MMI}
\end{table}

  \begin{table}[!htb]
  \begin{center} \footnotesize{
  \begin{tabular}{|l|c|}
  \hline
  Method & Accuracy ($\%$) \\
  \hline\hline
  LBP-TOP~\cite{Zhao:2007:DTR} & 68.13 \\
  HOG 3D~\cite{Klaser_aspatio-temporal2008} & 70.63 \\
  \hline
  AdaLBP~\cite{zhao2011oulucasia} & 73.54 \\
  Atlases~\cite{guo2012dynamic} & 75.52 \\
  STM-ExpLet~\cite{liu2014learning} & 74.59 \\
  DTAGN-Joint~\cite{jung2015joint} & 81.46 \\
  traj. on S+(2; n)~\cite{kacem2017novel} & 83.13\\
  \hline
  FN2EN ~\cite{ding2017facenet2expnet} & 87.71\\
  \hline
  CNN (baseline) & 83.96 \\
  Ours ($\textrm{GCNet}_{S0R0}$) & \textbf{84.65}\\
  Ours ($\textrm{GCNet}_{S1R0}$) & \textbf{86.39}\\
  Ours ($\textrm{GCNet}_{S0R1}$) & \textbf{85.83}\\
  Ours ($\textrm{GCNet}_{S1R1}$) & \textbf{86.11}\\
  \hline
\end{tabular} }
\end{center}
\vspace*{-1.0em}
\caption{Averaged recognition accuracy ($\%$) on the Oulu-CASIA VIS database, 6 expressions.}
\label{table:OULU}
\end{table}

In the CK+ database, seven expressions and a neutral image are included. We conducted experiments for seven expressions as well as eight expressions (seven expressions and a neutral face). For the seven expressions cases shown in Table~\ref{table:CK+seven}, the proposed methods ($\textrm{GCNet}_{S0R0}$, $\textrm{GCNet}_{S1R0}$, $\textrm{GCNet}_{S0R1}$, and $\textrm{GCNet}_{S1R1}$) show a better recognition performance than that of all compared state-of-the-arts including hand-craft feature based methods (LBP-TOP~\cite{Zhao:2007:DTR} and HOG 3D~\cite{Klaser_aspatio-temporal2008}), CNN-based methods (3D-CNN~\cite{liu2014learning}, 3D-CNN-DAP~\cite{liu2014learning}, and DTAGN-Joint~\cite{jung2015joint}), and video-based methods (MSR~\cite{conf/iccvw/PtuchaTS11}, TMS~\cite{jain2011facial}, STM-ExpLet~\cite{liu2014learning}, DTAGN-Joint~\cite{jung2015joint}, and  traj. on S+(2; n)~\cite{kacem2017novel}). For cases of the eight expressions shown in Table~\ref{table:CK+eight}, the proposed methods ($\textrm{GCNet}_{S0R0}$, $\textrm{GCNet}_{S1R0}$, $\textrm{GCNet}_{S0R1}$, and $\textrm{GCNet}_{S1R1}$) show a better recognition performance than the compared deep learning-based methods including FAU aware methods (AURF~\cite{liu2013aware}, AUDB~\cite{liu2015inspired}) and a CNN--based method (Zero-bias CNN+AD~\cite{khorrami2015deep}). When a loss function of contrastive metric learning is eliminated ($\textrm{GCNet}_{S0R0}$ and $\textrm{GCNet}_{S0R1}$), we observed that the performance is degraded than that with a contrastive loss ($\textrm{GCNet}_{S1R0}$ and $\textrm{GCNet}_{S1R1}$) on the CK+ database.

In the MMI database, similar to the case of the CK+ database, the proposed methods show a higher or comparable accuracy value than that of the state-of-the-arts including CNN-based methods (3D-CNN-DAP~\cite{liu2014learning} and DTAGN-Joint~\cite{jung2015joint}) and video-based methods (STM-ExpLet~\cite{liu2014learning}, DTAGN-Joint~\cite{jung2015joint}, and  traj. on S+(2; n)~\cite{kacem2017novel}) as shown in Table~\ref{table:MMI}. The methods (STM-ExpLet~\cite{liu2014learning}, DTAGN-Joint~\cite{jung2015joint},  and  traj. on S+(2; n)~\cite{kacem2017novel}) that acquire temporal information from multiple images show relatively higher accuracy performance than other methods. Even though the proposed methods show a better recognition performance than these compared methods, the recognition accuracy of the proposed methods on the MMI database is relatively less compared to that on other databases (CK+ and Oulu-CASIA VIS). Due to the large intra-identity variation of the MMI database, locally selected patch based method (CSPL~\cite{zhong2012learning},) shows a relatively better performance than other compared methods. 

In the Oulu-CASIA VIS database, the proposed methods show higher or comparable accuracy values with CNN-based methods (DTAGN-Joint~\cite{jung2015joint} and FN2EN~\cite{ding2017facenet2expnet}) and video-based methods (AdaLBP~\cite{zhao2011oulucasia}, Atlases~\cite{guo2012dynamic}, STM-ExpLet~\cite{liu2014learning}, DTAGN-Joint~\cite{jung2015joint},  traj. on S+(2; n)~\cite{kacem2017novel}), as shown in Table~\ref{table:OULU}. Our proposed networks include the smaller number of parameters, 3M parameters for $\textrm{GCNet}_{S1R1}$, than 11M parameters for the FN2EN~\cite{ding2017facenet2expnet}, with a comparable recognition performance.

\subsection{Qualitative analysis}

\subsubsection{Visualization of the response maps}
\begin{figure}[!hbt]
\begin{center}
\centering
\includegraphics[width=0.99\linewidth]{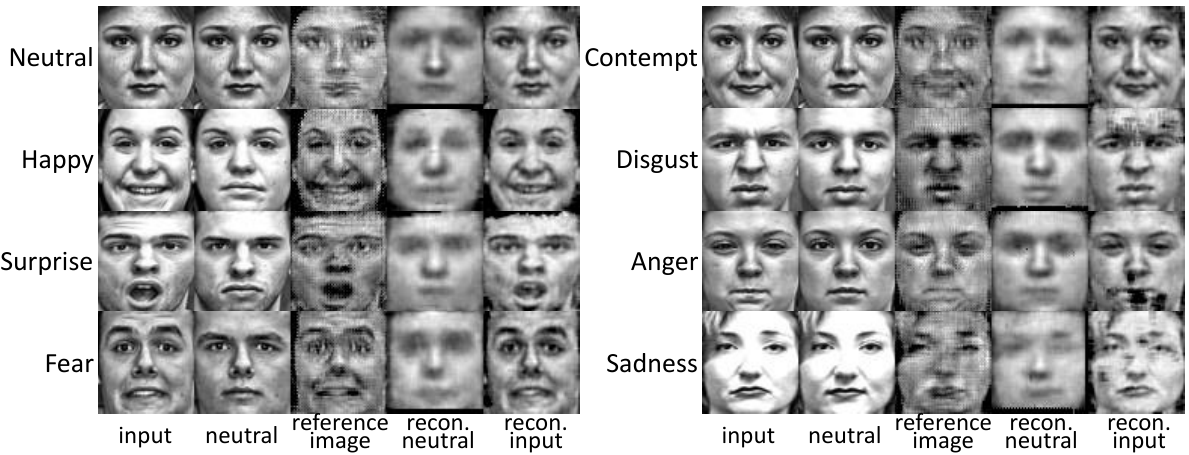}
\end{center}
\vspace{-.5em}
   \caption{Examples of generation and reconstruction results on the test data. The reconstructed images (recon. neutral and recon. input) are not necessary for the classification task in the test phase, but shown for a reference.  }
\label{fig:response_map_CK+}
\end{figure}

We observe the response maps resulted from generation and reconstruction layers of the proposed networks to understand what the networks have been conducted in the test phase. In Figure~\ref{fig:response_map_CK+}, a generated reference image, a reconstructed neutral image, and a reconstructed image of a given expression are shown. The generated reference image is affected by reconstruction and contrastive metric learning. Even though the reconstruction images do not affect contrastive representation  in the test phase, we show the images for a reference.

\subsubsection{Comparison of discriminative distributions}
\begin{figure}[!hbt]
\begin{center}
\hspace*{-1.0em}
   \subfloat[{Image Pixel data} ]{\includegraphics[width=0.36\linewidth]{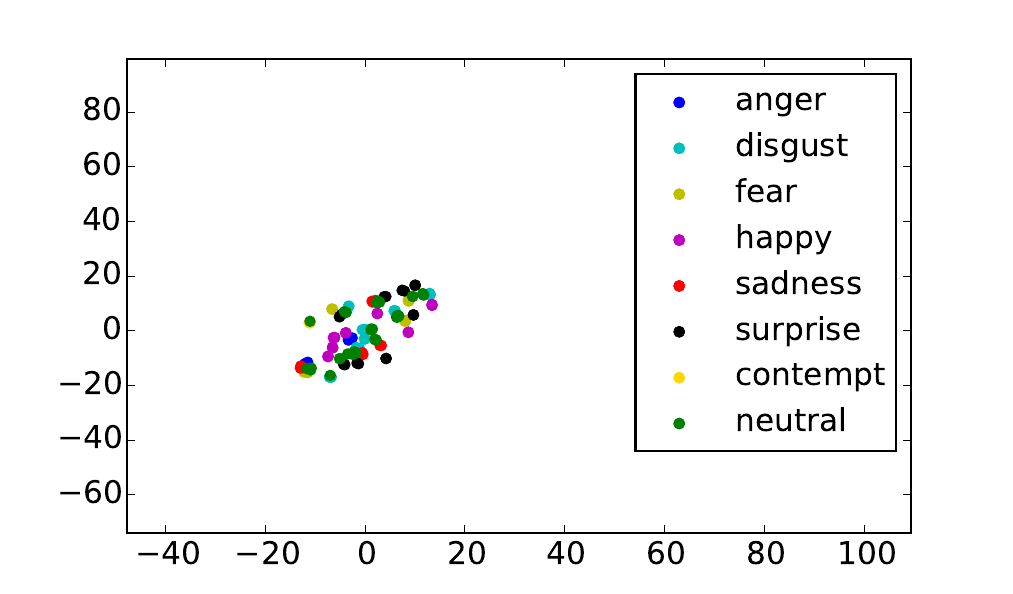}}
  \hspace*{-1.0em}
  \subfloat[{CNN (baseline) feature} ]{\includegraphics[width=0.36\linewidth]{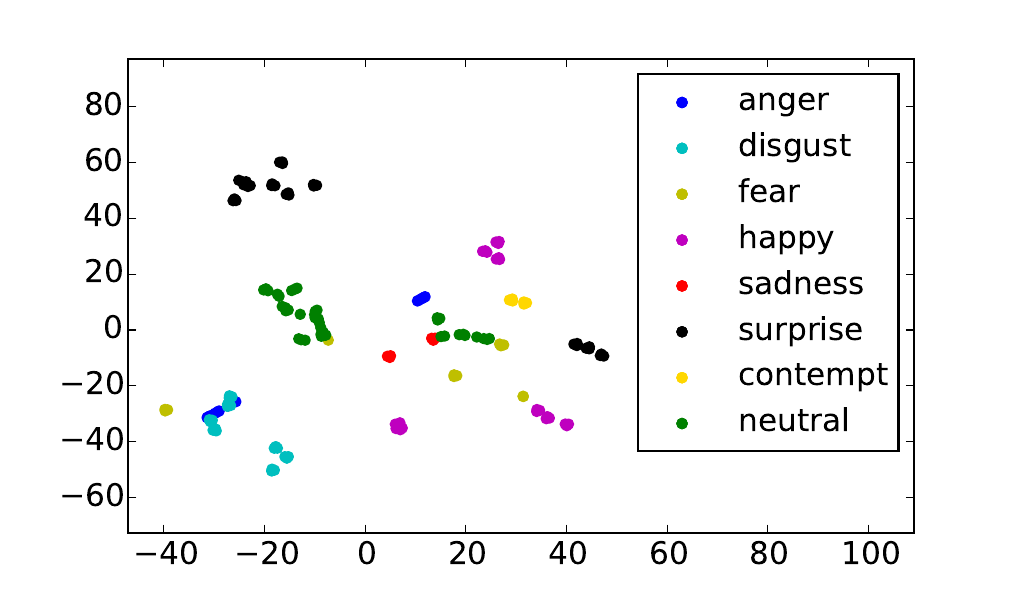}}
  \hspace*{-1.0em}
  \subfloat[{Ours ($\textrm{GCNet}_{S1R1}$)} ]{\includegraphics[width=0.36\linewidth]{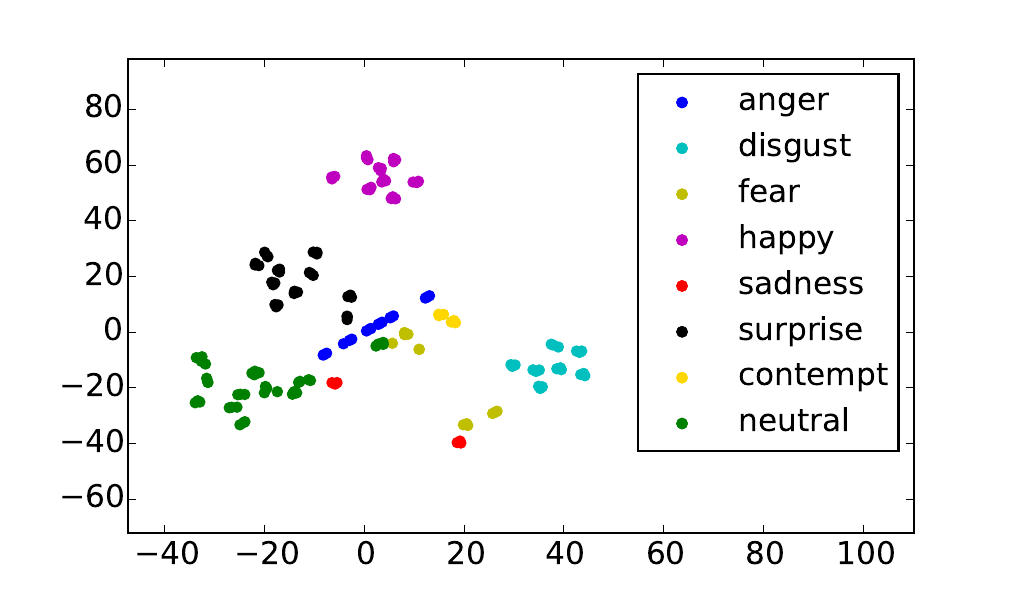}}
  \hspace*{-1.0em}
\end{center}
\vspace{-0.7em}
   \caption{Visualization of the extracted features using t-SNE: (a) a pixel value of the input images, (b) a feature vector of CNN (baseline), and (c) a feature vector of the proposed method ($\textrm{GCNet}_{S1R1}$).}
\label{fig:tSNE}
\end{figure}

To observe a discriminative distribution of the extracted features, we visualized the feature vectors from the first layer of the fully-connected layers of the proposed networks and the CNN (baseline). We visualize the 384 dimensional feature vectors using t-SNE~\cite{maaten2008visualizing}. As shown in Figure~\ref{fig:tSNE}, the feature points of original images are scattered within a narrow region. The point distribution of the CNN (baseline) forms partially overlapped clusters. The proposed network features are appropriately clustered to discriminate individual expression further.

\subsubsection{Distributions of features along gradual expression changes}
\begin{figure}[!hbt]
\begin{center}
\centering
 \vspace*{-1.0em}
 \hspace*{-1.0em}
  \subfloat[{t-SNE of CNN (baseline) } ]{\includegraphics[width=0.48\linewidth]{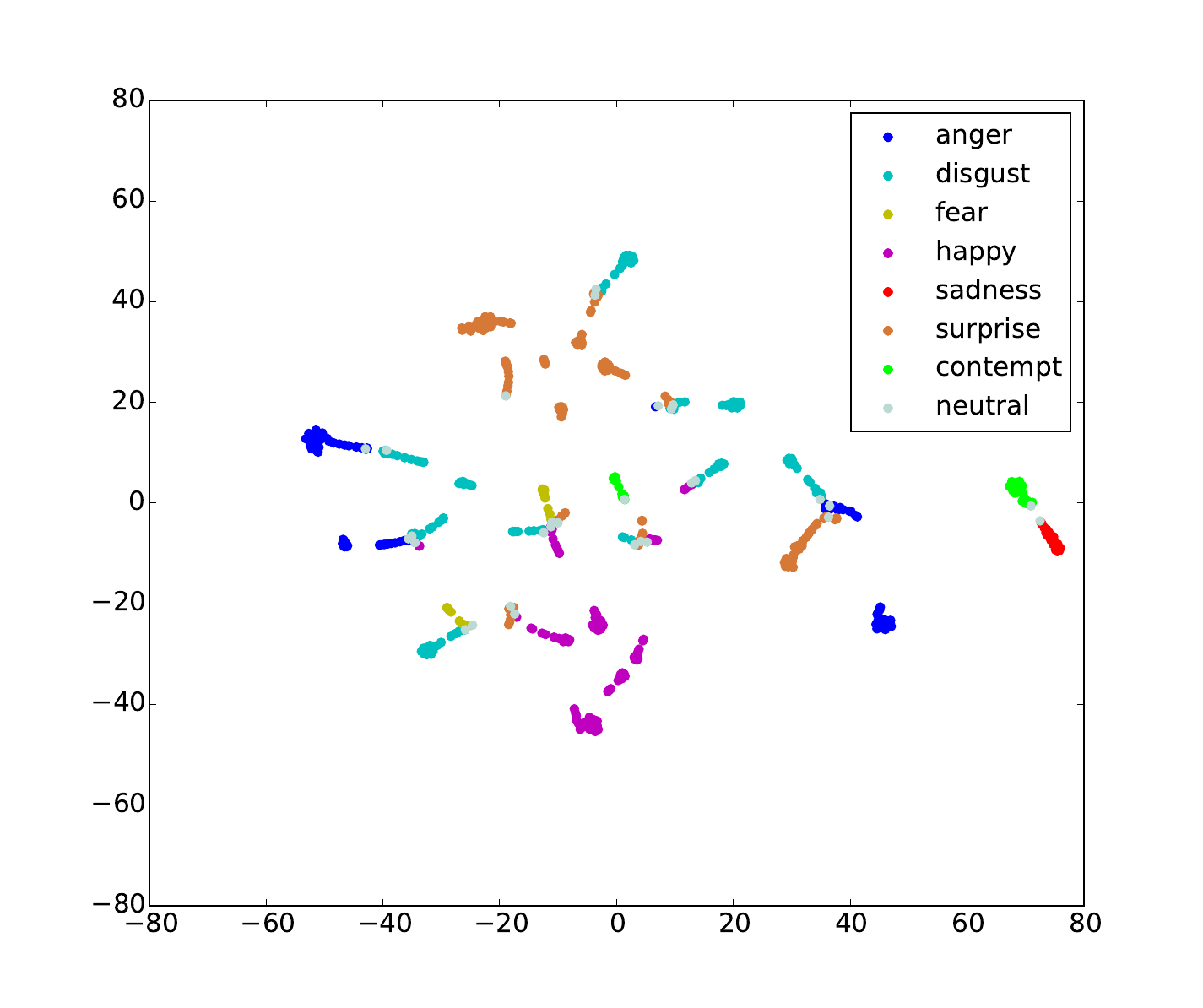}}
  \hspace*{-1.0em}
  \subfloat[{t-SNE of Ours ($\textrm{GCNet}_{S1R1}$)} ]{\includegraphics[width=0.48\linewidth]{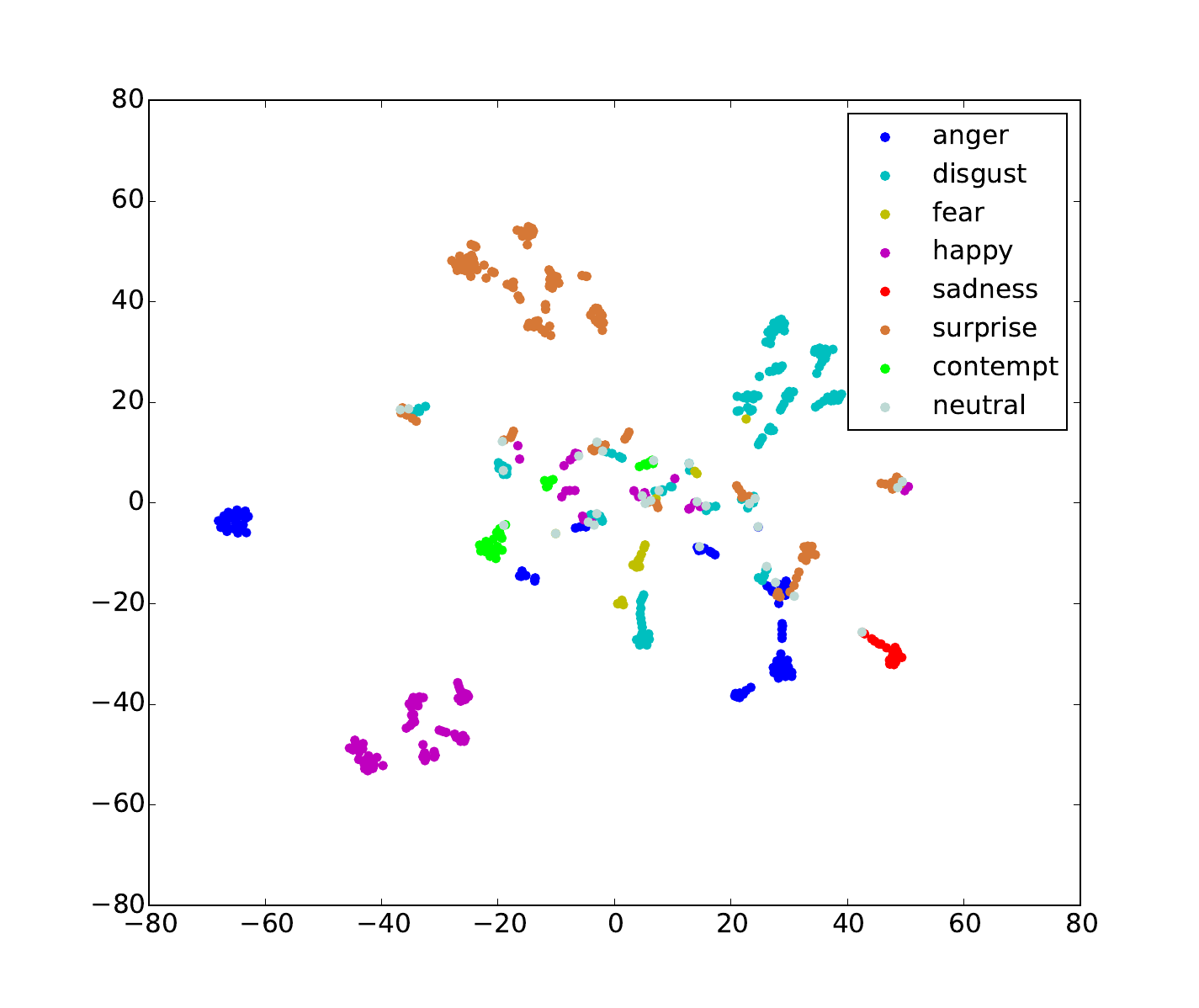}}
  \par
  \subfloat[{Image mappings on CNN (baseline)} ]{\includegraphics[width=0.48\linewidth]{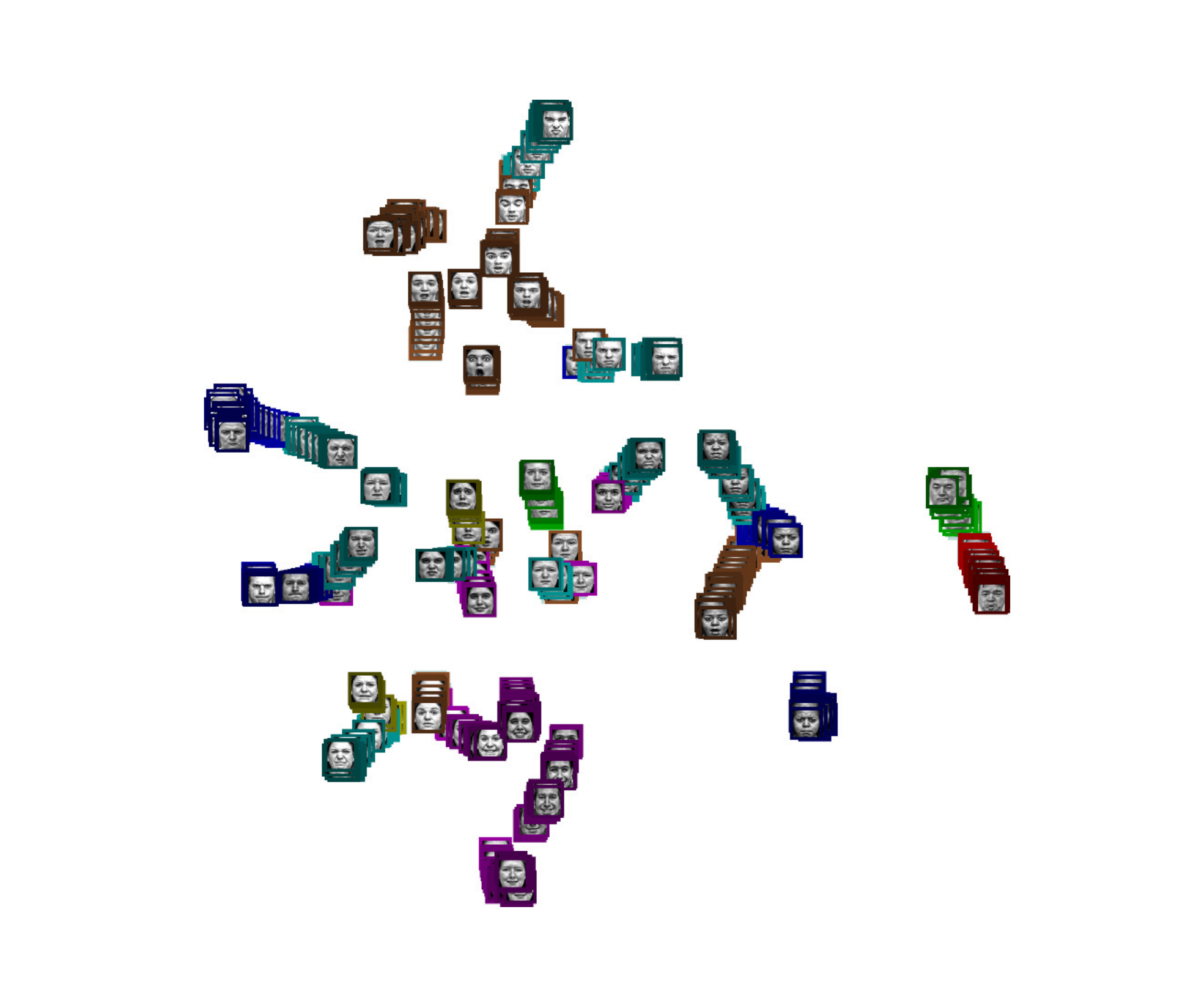}}
  \subfloat[{Image mappings on Ours ($\textrm{GCNet}_{S1R1}$)} ]{\includegraphics[width=0.48\linewidth]{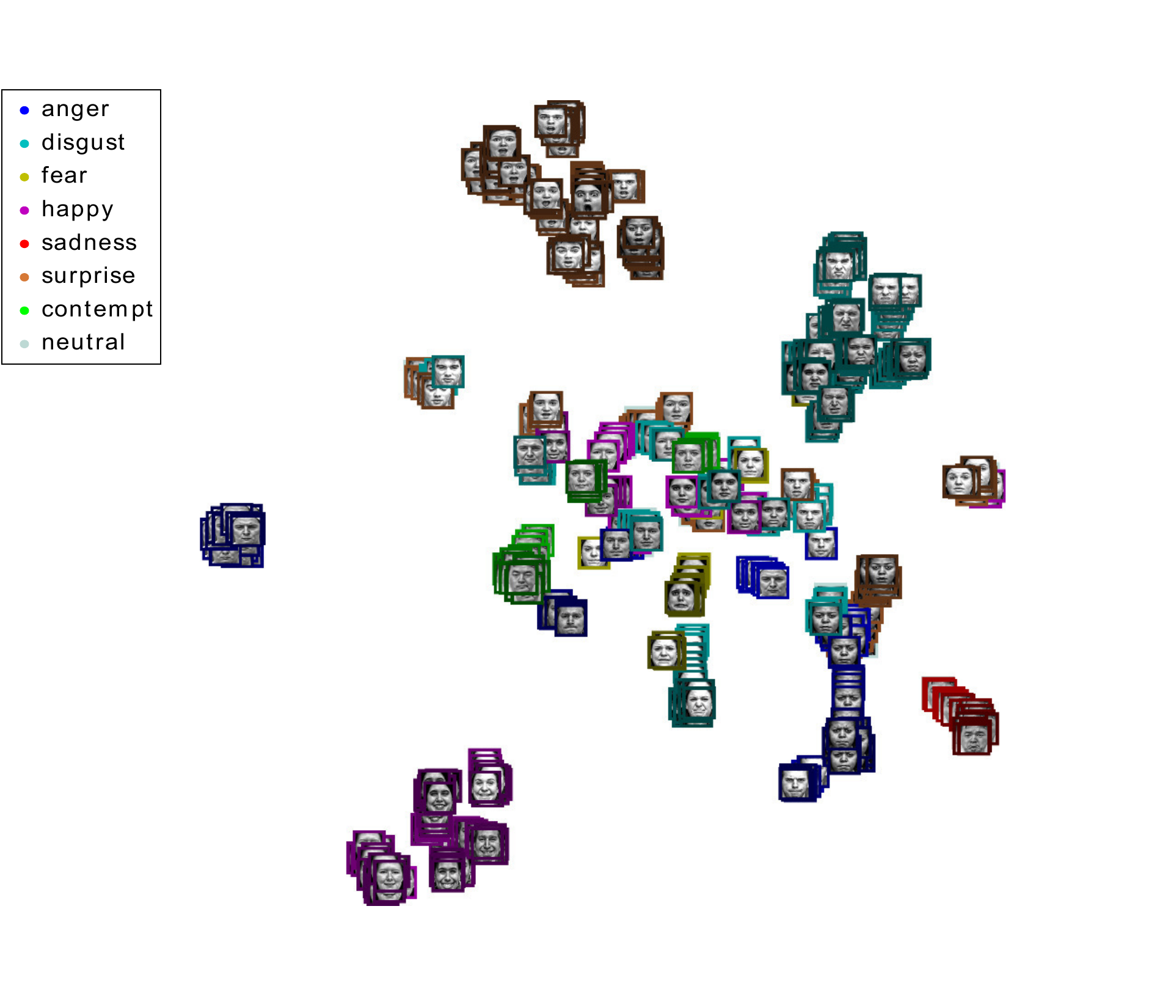}}
\end{center}
   \caption{Visualization using t-SNE of the extracted features from tested image sequences (one-fold selected from CK+ database arbitrarily): using a feature vector of (a) CNN (baseline) and (b) our proposed method, corresponding facial images to points in the t-SNE distribution (c) the points of (a) and (d) the points of (b) are visualized. In (c) and (d), note that color painted around each image are labelled with a corresponding expression label. The lighter color indicates the smaller expressiveness.  Our proposed method shows that the points shown same expression are distributed in a narrow region.  In the CNN (baseline), the points shown different expressions with a small degree of expressiveness are distributed closely.}
\label{fig:tSNE_sequence}
\end{figure}

 We observe the distributions of the feature vectors extracted from the sequential images (one sequence includes images from neutral to expression) using the t-SNE \cite{maaten2008visualizing} in Figure~\ref{fig:tSNE_sequence}. In our proposed method, the feature vectors from the same expression tend to be distributed in the same cluster. Unclustered samples are mostly belonged to neutral-like (less-expressive) images.
 In the CNN (baseline), the samples are distributed closely with different expressions. It means that indistinctive representation is extracted among different level of expressive images.


\subsection{Effects of the generative reference image with the generator networks}
\begin{figure}[hbt]
\begin{center}
\hspace*{-0.8em} 
\subfloat[$\textrm{GCNet}_{S1R1}$]{\includegraphics[width=0.29\linewidth]{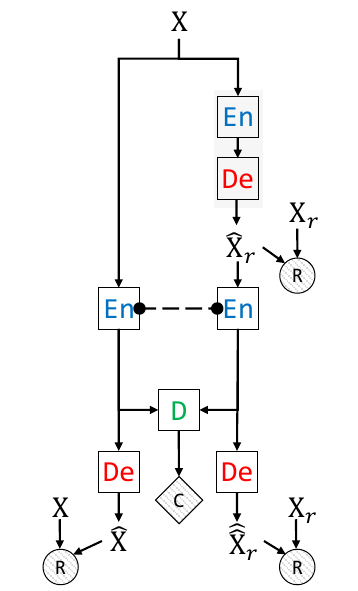} }
\hspace*{-0.8em} 
\subfloat[$\textrm{GCNet}_{S1R1}$\newline$+$ConvVAEs]{\includegraphics[width=0.29\linewidth]{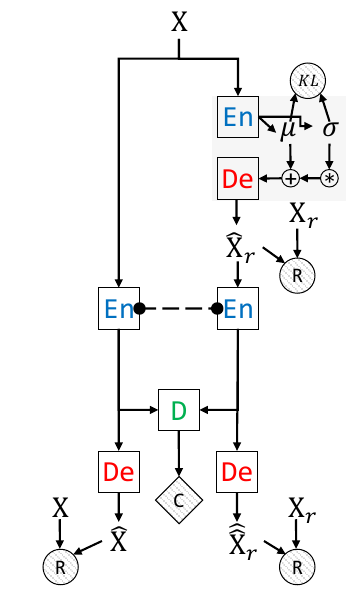} }
\hspace*{-0.8em} 
\subfloat[$\textrm{GCNet}_{S1R1}$\newline$+$DCGANs]{\includegraphics[width=0.29\linewidth]{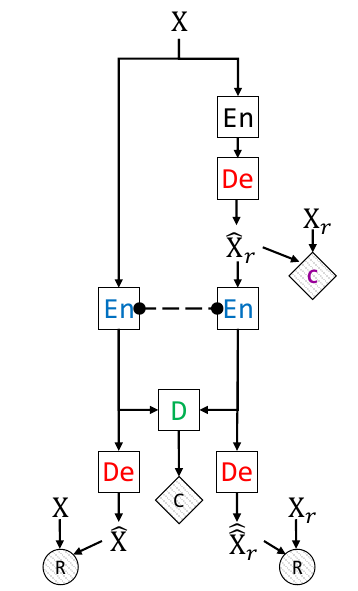} }%
\hspace*{-0.8em} 
\subfloat{\includegraphics[width=0.25\linewidth]{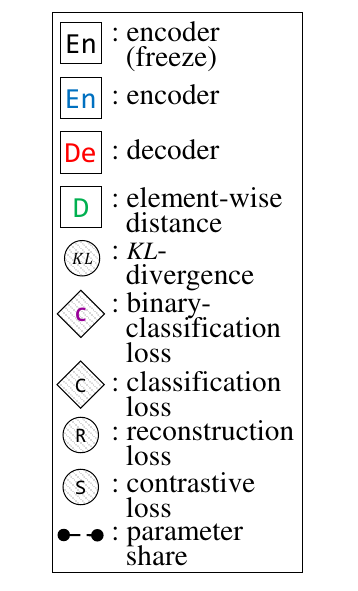} } %
\end{center}
\vspace*{-1.5em}
   \caption{Architecture overviews of our proposed networks (GCNet) and combinations with other generative networks:
   (a) our proposed method $\textrm{GCNet}_{S1R1}$,
   (b) the generator networks of $\textrm{GCNet}_{S1R1}$ are replace with VAEs, and
   (c) the generator networks of $\textrm{GCNet}_{S1R1}$ are replace with GANs.}
\label{fig:gcnet_variants}
\end{figure}

The generator in our proposed methods can adopt other generative models for estimating the reference image. In this section, we evaluate the proposed network ($\textrm{GCNet}_{S1R1}$) with two representative baseline generative models such as the Variational Auto-Encoders (VAEs)~\cite{KingmaW13} and Generative Adversarial Networks (GANs)~\cite{Goodfellow:NIPS2014, RadfordMC15}.
 In Figure~\ref{fig:gcnet_variants}, we show modified architectures that the generator networks of the proposed networks are replaced with those generative networks.
We utilize the \textit{convolutional} layers to the generative networks (VAEs and GANs), as \textit{convolutional} encoder-decoder networks (denoted by ``Conv.AEs") is adopted in the proposed network. After adopting the convolutional layers, we named Conv.VAEs for the VAEs method and DCGANs for the GANs respectively. We follow the network structure in the literature of DCGANs~\cite{RadfordMC15} for GANs\footnote{As GAN is more effectively implemented in the PyTorch environment than in Caffe, our experiments are based on the open PyTorch codes. \url{https://github.com/pytorch/examples/blob/master/dcgan/main.py}}. For VAEs\footnote{Fully connected layers based VAEs from \url{https://github.com/pytorch/examples/blob/master/vae/main.py} are replaced with convolutional layers. The number of channels of the last conv. layer of the encoder is set to two times of the dimensionality of a latent vector.}, we adopt the layer structure of the discriminator of the DCGANs for the the encoders, and the generator of DCGANs for the decoders respectively. All evaluations in this section are conducted using the \textsf{PyTorch} framework on the CK+ dataset (9-folds for training and 1-fold for test). The parameters of the networks are learned without the pre-training process.


\begin{figure*}[hbt]
\begin{center}
 \hspace*{-0.2em} 
  \subfloat[$\textrm{GCNet}_{S1R1}$]{\includegraphics[width=0.33\linewidth]{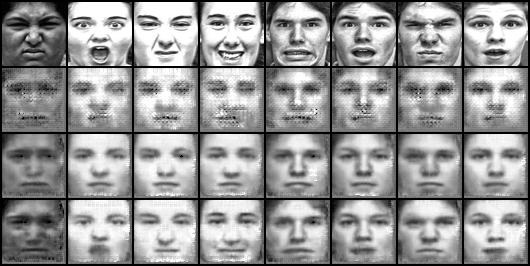}}
  \hspace*{-0.2em} 
  \subfloat[$\textrm{GCNet}_{S1R1}$$+$Conv.VAEs]{\includegraphics[width=0.33\linewidth]{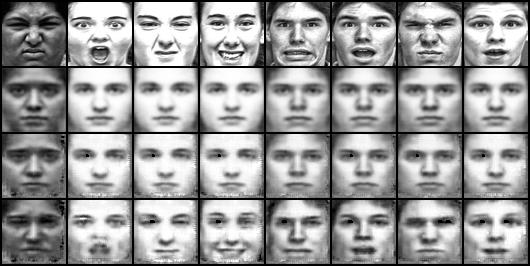}}
  \hspace*{-0.2em} 
\subfloat[$\textrm{GCNet}_{S1R1}$$+$DCGANs]{\includegraphics[width=0.33\linewidth]{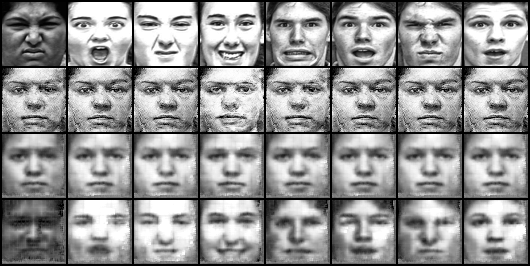}} \\
\subfloat[ $\textrm{GCNet}_{S0R1}$ without a classification task]{\includegraphics[width=0.33\linewidth]{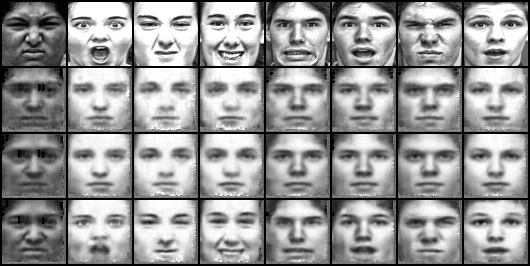}}
  \subfloat[Recognition accuracy]{\includegraphics[width=0.33\linewidth]{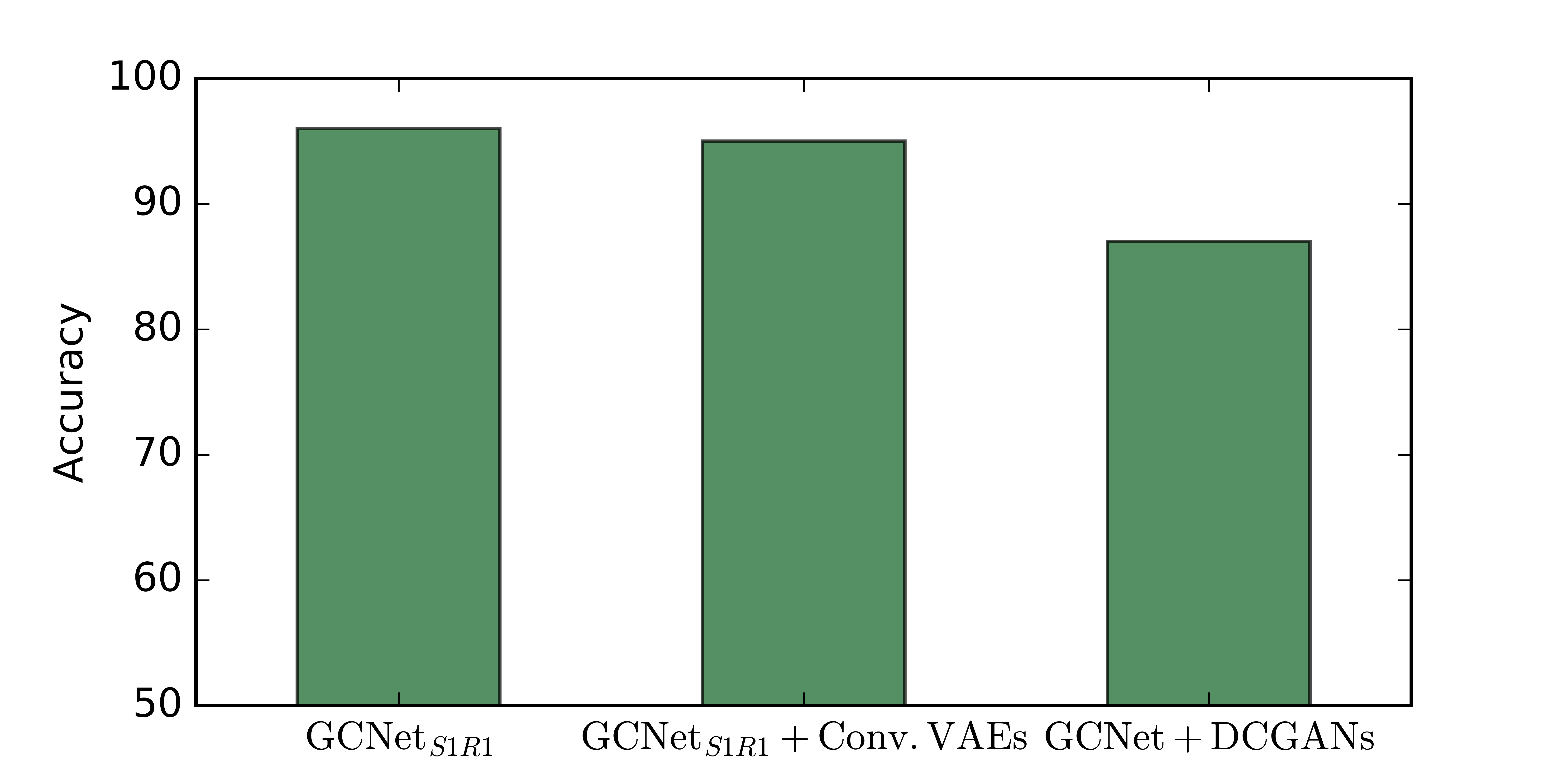}}
\end{center}
\vspace{-0.5em}
   \caption{The generated images (a generative reference image in the 2nd row, a reconstructed image of the generative reference image in the 3rd row, and a reconstructed input image in the 4th row) of the query image in the first row are shown: (a) the proposed $\textrm{GCNet}_{S1R1}$, (b) Conv.VAEs based $\textrm{GCNet}_{S1R1}$, (c) DCGANs based $\textrm{GCNet}_{S1R1}$, and(a) $\textrm{GCNet}_{S0R1}$ without classification layers. In (e) the corresponding accuracy values to (a), (b), and (c) are shown. }
\label{fig:GCnet_variants_generated images}
\end{figure*}

\subsubsection{Generation quality and discrimination performance}
We observe how the generative networks affect the image quality and accuracy performance. In Figure~\ref{fig:GCnet_variants_generated images}~(a), (b), and (c), the generated images (a generative reference image (2nd row), a reconstructed image of the generative reference image (3rd row), and a reconstructed image of the input real image (4th row)) from the input real image (1st row) are shown. The Conv.VAEs based $\textrm{GCNet}_{S1R1}$ shows a comparable recognition performance with the proposed Conv.AEs based networks  as shown in Figure~\ref{fig:GCnet_variants_generated images}~(e). Even though Conv.VAEs based networks at (b) show clearer generative reference images than Conv.AEs based networks at (a), an individual's identity factor seems slightly less preserved by the Conv.VAEs based method than that by the Conv.AEs based method. This might be one of reasons why the VAEs based method shows slightly lower accuracy than the proposed Conv.AEs based one. The DCGANs based networks at (c) fails to preserve the identity. This is might be because the identity-preserving loss is not contained.

\subsubsection{Number of parameters}
In Figure~\ref{fig:GCnet_variants_generated images}, if the larger number of parameters is utilized in the generator, the better generated image is shown with a similar recognition performance (please refer to 1.7 M parameters for Conv.AEs (a) $vs.$  9.6 M for Conv.VAEs (b)). The better visual quality on human eyes might be not necessary for the better recognition performance~\cite{Nguyen_2015_CVPR}, in the proposed networks.  

As shown in  Figure~\ref{fig:GCnet_variants_generated images}~(d), the generative reference image without a contrastive loss and a discriminative loss is more clearly generated than that in (a).
The number of parameters at (a) might be not enough to generate a good quality image and conduct the contrastive and discriminative learning at the same time comparing to the VAE case.

\subsection{Experiments using more unconstrained data}
 In this section, we conducted an experiment using ``in-the-wild'' databases such as RAF~\cite{Li_2017_CVPR}, FER2013~\cite{GOODFELLOW201559}, and SFEW~\cite{Emotiw2015} databases. The images of RAF and FER2013 are collected from the internet search engine. The images of SFEW are selected from short video clips where their expression label per each frame is annotated using an automatic approach. More details refer to the corresponding literatures \cite{Li_2017_CVPR,GOODFELLOW201559, Emotiw2015}.

 As shown in Table~\ref{table:cross-db}, our proposed method shows a comparable or outperformed recognition performance to the compared methods. As a pair of images (e.g. neutral and expression) is unavailable during training in the unconstrained database, we experimented with a \textit{cross-database} setting where training and test databases are different.

\begin{table}[!hbt]
  \begin{center}\scriptsize{
  \begin{tabular}{|l|c|c|l| }
  \hline
  Method & Acc. ($\%$) & Test &Training \\
  \hline\hline
  Inception-CNN~\cite{fer_wacv2016}                       & 34.00              & \multirow{4}{*}{\textbf{FER2013}~\cite{GOODFELLOW201559}} & CK+$+$MMI$+$5 DBs\\
  \cline{1-2}\cline{4-4}
  Ours ($\textrm{GCNet}_{S1R1}$) & \textbf{40.19}  &   & CK+\\
  \cline{1-2}\cline{4-4}
  Ours ($\textrm{GCNet}_{S1R1}$) & \textbf{41.71}  &   & CK+$+$OuluCASIA\\
  \cline{1-2}\cline{4-4}
  Ours ($\textrm{GCNet}_{S1R1}$) & \textbf{45.43}  &   & CK+$+$MMI$+$OuluCASIA\\
  \hline\hline
  traj. on $\mathcal{S}^{+}$~\cite{kacem2017novel}   & 39.94*  & \textbf{AFEW}~\cite{Emotiw2015} &  \textbf{AFEW}~\cite{Emotiw2015} \\
  \hline
  FN2EN~\cite{ding2017facenet2expnet}                         & 48.19*  & \multirow{4}{*}{\textbf{SFEW}~\cite{Emotiw2015}} &  \multirow{2}{*}{\textbf{SFEW}~\cite{Emotiw2015}} \\
  \cline{1-2}
  DLP-CNN~\cite{Li_2017_CVPR}                                     & 51.05*  &             &   \\
  \cline{1-2} \cline{4-4}
  Inception-CNN~\cite{fer_wacv2016}                       & 39.80              &  & CK+$+$MMI$+$5 DBs\\
  \cline{1-2} \cline{4-4}
  Ours ($\textrm{GCNet}_{S1R1}$)  & \textbf{35.57}  &                                            &  CK+$+$MMI$+$OuluCASIA\\
  \hline
\hline
  DLP-CNN+SVM~\cite{Li_2017_CVPR}   & 74.2*                  & \multirow{5}{*}{\textbf{RAF}~\cite{Li_2017_CVPR}}& \textbf{RAF}~\cite{Li_2017_CVPR}\\
  \cline{1-2} \cline{4-4}
  HOG+SVM~\cite{Li_2017_CVPR}  &   39                &     & \multirow{2}{*}{CK+}\\
  \cline{1-2}
  Ours ($\textrm{GCNet}_{S1R1}$) & \textbf{45.44} &   & \\
  \cline{1-2} \cline{4-4}
  Ours ($\textrm{GCNet}_{S1R1}$) & \textbf{49.80} &   &CK+$+$OuluCASIA \\
  \cline{1-2} \cline{4-4}
  Ours ($\textrm{GCNet}_{S1R1}$) & \textbf{49.74} &   & CK+$+$MMI$+$OuluCASIA \\
  \hline
  \hline
  \multicolumn{4}{c}{*same database for training and test.}  \\
  \end{tabular}
  }
\end{center}
\vspace{-0.3em}
\caption{The recognition accuracy ($\%$) on the in-the-wild databases with cross-database settings. 7 expressions. AEFW/SEFW: validation set.}
\label{table:cross-db}
\vspace{-1.0em}
\end{table}

\section{Conclusions}
In this paper, we proposed the facial expression recognition method based on contrastive representation learning. The contrastive representation is calculated in the embedding layers of deep networks by comparing a given image with a reference image. The reference image is generated by deep generative networks. The contrastive representation in the latent space is provided as the input to the final classification layers which conducts the expression recognition. Our proposed approach is useful especially if an expressive depth of an emotional face is varied among individuals, expressions, or situations. In our proposed networks, we attempted to disentangle a facial expressive factor directly. Disentangling of expression is conducted in two steps: 1) learning of the reference face by a generator network and 2) learning of the contrastive representation with a combination of contrastive and reconstruction objectives. The generative, contrastive, and discriminative learning is conducted in the end-to-end deep networks at the same time. Extensive experiments were conducted on three face expression databases that are publicly available and widely adopted in the literature. The proposed method outperforms the known state-of-the arts, including both single image and multiple-image based methods. This study could be extended to effectively detect and recognize small changes of facial expressions from sequential images. We will replace the current generator in the proposed networks with recent sophisticated generative networks to observe their effects in future.

\appendices
\section*{Appendix}
In this appendix, we present additional empirical results: 1) evaluations of the generative networks with respect to reconstruction and disentangle performances and 2) evaluations of recognition accuracy along with different weight values for contrastive learning loss.

 \begin{figure}[bt]
\begin{center}
\vspace*{-.5em}
  \subfloat[Reconstruction by Conv.VAEs]{\includegraphics[width=0.74\linewidth]{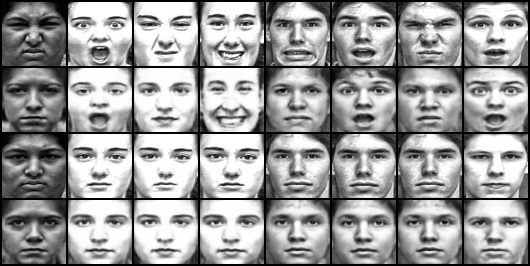}}\\
  \vspace*{-.5em}
\subfloat[Reconstruction by DCGANs]{\includegraphics[width=0.74\linewidth]{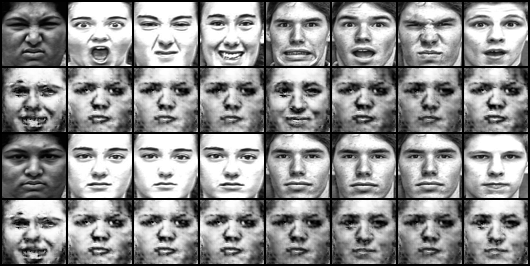}}\\
  \vspace*{-.5em}
\end{center}
\vspace{-0.7em}
   \caption{Reconstructed expression images (2nd row) and reconstructed neutral images (4th row) for the original expression images (1st row) and the neutral images (3rd row) by using (a) Conv.VAEs and (b) DCGANs respectively.}
\label{fig:reconstruction}
\end{figure}

\begin{figure*}[hbt]
\begin{center}
 \hspace*{-0.2em}
  \subfloat[Disentangled by Conv.VAEs]{\includegraphics[width=0.33\linewidth]{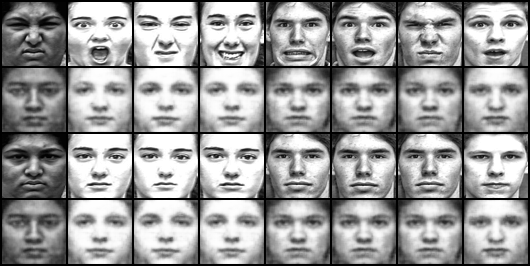}}
 \hspace*{-0.2em}
  \subfloat[Disentangled by DCGANs]{\includegraphics[width=0.33\linewidth]{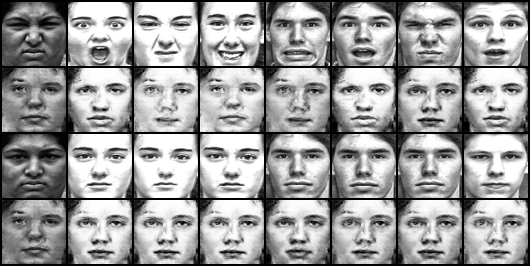}}
 \hspace*{-0.2em}
  \subfloat[Disentangled by Conv.AEs]{\includegraphics[width=0.33\linewidth]{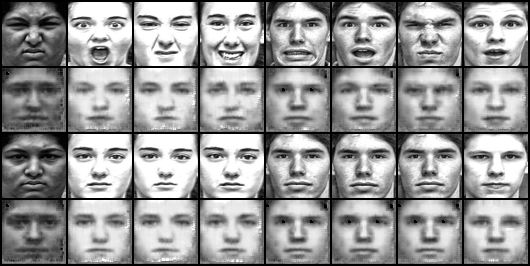}}
\end{center}
\vspace{-.5em}
   \caption{Disentangled expression images (2nd row) and disentangled neutral images (4th row, neutral reconstruction) for the original expression images (1st row) and the neutral images (3rd row) by using (a) Conv.VAEs, (b) DCGANs, and (c) Conv.AEs respectively. }
\label{fig:disentangle}
\end{figure*}

\begin{figure}[hbt]
\begin{center}
\centering
 \vspace*{-1.0em}
 \subfloat[Along different weights of a contrastive loss]{\includegraphics[width=0.75\linewidth]{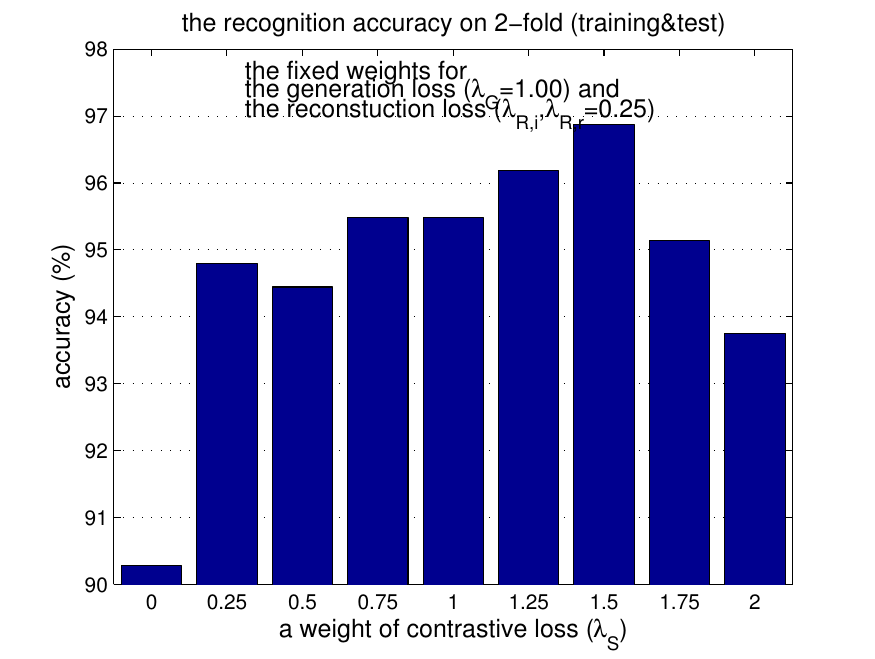}}
\end{center}
 \vspace*{-1.0em}
   \caption{Accuracy ($\%$)  observation along the different weights for the contrastive loss function with the fixed weights for both the generation loss and the   reconstruction loss.}
\label{fig:parameter_contrastive}
\end{figure}

\subsection{Reconstructed and disentangled faces using the generative networks}
For a better understanding of the results in the experimental sections, we show the reconstruction and disentanglement performances of expression images using the Conv.VAEs and DCGANs without adding any additional networks. As the original DCGANs are not designed to reconstruct the images, we added an encoder layers to map an image (in the data space, higher dimensional) into an embedding vector (in the latent space, lower dimensional) using the
fully connected layer without training (a random projection transformation\footnote{by the Johnson-Lindenstrauss lemma~\cite{johnson84extensionslipschitz},  a random projection preserves all pairwise distances between the points which are in the subspace of the higher-dimensional Euclidean space with a high probability.}): $\mathbb{R}^{h\times w} \rightarrow \mathbb{R}^{p}$.

In Figure~\ref{fig:reconstruction}, we show the reconstruction performance of  (a) Conv.VAEs and (b) DCGANs. The real expression image (1st row) and real neutral image (3rd row) are targets to the reconstruction (2nd and 4th rows respectively). In Figure~\ref{fig:reconstruction}, the Conv.VAEs shows a better reconstruction performance than DCGANs. The DCGANs show almost random generative face images (2nd row). In order to generate an image with preserving the expression and the identity, additional loss functions designed with supervised settings consists of expression and identity labels might be needed.

In Figure~\ref{fig:disentangle}, we observe the performance to disentangle an expression factor (from an expressive face to a less-expressive face). The real expression image (1st row) and real neutral image (3rd row) are transformed into disentangled images (2nd and 4th rows respectively). As shown in Figure~\ref{fig:disentangle}~(a) and (b), both Conv.VAEs and DCGANs can disentangle the expressive factor. However, the identity factor is also disentangled, even though a pair of a given image and a target image with the same identity is used during training in this experiment. The Conv.AE shows relatively better performance to preserve the identity information than the compared methods as shown in Figure~\ref{fig:disentangle}~(c).

\subsection{Effects of the weight for the loss function}
As shown in Figure~\ref{fig:parameter_contrastive}, as a weight value of the contrastive loss increases, the accuracy value increases till a certain degree. In these evaluations, 2-fold validation is adopted as 10-fold cross-validation requires too many evaluation cases due to extensive hyper-parameter settings. The 1st and 2nd folds among 10 folds are selected for a test set, and the remained folds are for a training set.

\section*{Acknowledgment}
The authors would like to thank members of \textit{Face Intelligence} project for their kind helps and the supercomputer center for their GPU-server supports at SAIT.

\ifCLASSOPTIONcaptionsoff
  \newpage
\fi


\begin{thebibliography}{10}
\providecommand{\url}[1]{#1}
\csname url@samestyle\endcsname
\providecommand{\newblock}{\relax}
\providecommand{\bibinfo}[2]{#2}
\providecommand{\BIBentrySTDinterwordspacing}{\spaceskip=0pt\relax}
\providecommand{\BIBentryALTinterwordstretchfactor}{4}
\providecommand{\BIBentryALTinterwordspacing}{\spaceskip=\fontdimen2\font plus
\BIBentryALTinterwordstretchfactor\fontdimen3\font minus
  \fontdimen4\font\relax}
\providecommand{\BIBforeignlanguage}[2]{{%
\expandafter\ifx\csname l@#1\endcsname\relax
\typeout{** WARNING: IEEEtran.bst: No hyphenation pattern has been}%
\typeout{** loaded for the language `#1'. Using the pattern for}%
\typeout{** the default language instead.}%
\else
\language=\csname l@#1\endcsname
\fi
#2}}
\providecommand{\BIBdecl}{\relax}
\BIBdecl

\bibitem{haxby2000distributed}
J.~V. Haxby, E.~A. Hoffman, and M.~I. Gobbini, ``The distributed human neural
  system for face perception,'' \emph{Trends in cognitive sciences}, vol.~4,
  no.~6, pp. 223--233, 2000.

\bibitem{bruce1986understanding}
V.~Bruce and A.~Young, ``Understanding face recognition,'' \emph{British
  journal of psychology}, vol.~77, no.~3, pp. 305--327, 1986.

\bibitem{calder2005understanding}
A.~J. Calder and A.~W. Young, ``Understanding the recognition of facial
  identity and facial expression,'' \emph{Nature Reviews Neuroscience}, vol.~6,
  no.~8, pp. 641--651, 2005.

\bibitem{Bengio:2013:RLR}
Y.~Bengio, A.~Courville, and P.~Vincent, ``Representation learning: A review
  and new perspectives,'' \emph{IEEE Trans. Pattern Anal. Mach. Intell.},
  vol.~35, no.~8, pp. 1798--1828, Aug. 2013.

\bibitem{liu2014learning}
M.~Liu, S.~Shan, R.~Wang, and X.~Chen, ``Learning expressionlets on
  spatio-temporal manifold for dynamic facial expression recognition,'' in
  \emph{Proceedings of the IEEE Conference on Computer Vision and Pattern
  Recognition}, 2014, pp. 1749--1756.

\bibitem{bartlett2005recognizing}
M.~S. Bartlett, G.~Littlewort, M.~Frank, C.~Lainscsek, I.~Fasel, and
  J.~Movellan, ``Recognizing facial expression: machine learning and
  application to spontaneous behavior,'' in \emph{2005 IEEE Computer Society
  Conference on Computer Vision and Pattern Recognition (CVPR'05)},
  vol.~2.\hskip 1em plus 0.5em minus 0.4em\relax IEEE, 2005, pp. 568--573.

\bibitem{Zhao:2007:DTR}
G.~Zhao and M.~Pietikainen, ``Dynamic texture recognition using local binary
  patterns with an application to facial expressions,'' \emph{IEEE Trans.
  Pattern Anal. Mach. Intell.}, vol.~29, no.~6, pp. 915--928, Jun. 2007.

\bibitem{liu2013aware}
M.~Liu, S.~Li, S.~Shan, and X.~Chen, ``Au-aware deep networks for facial
  expression recognition,'' in \emph{Automatic Face and Gesture Recognition
  (FG), 2013 10th IEEE International Conference and Workshops on}.\hskip 1em
  plus 0.5em minus 0.4em\relax IEEE, 2013, pp. 1--6.

\bibitem{liu2015inspired}
------, ``Au-inspired deep networks for facial expression feature learning,''
  \emph{Neurocomputing}, vol. 159, pp. 126--136, 2015.

\bibitem{ekman:friesen:1978}
P.~Ekman and W.~Friesen, \emph{{Facial Action Coding System: A Technique for
  the Measurement of Facial Movement.}}\hskip 1em plus 0.5em minus 0.4em\relax
  Palo Alto: Consulting Psychologists Press, 1978.

\bibitem{kacem2017novel}
A.~Kacem, M.~Daoudi, B.~Amor, and J.~C. Alvarez-Paiva, ``A novel space-time
  representation on the positive semidefinite cone for facial expression
  recognition,'' in \emph{Proceedings of the IEEE Conference on Computer Vision
  and Pattern Recognition}, 2017, pp. 3180--3189.

\bibitem{jung2015joint}
H.~Jung, S.~Lee, J.~Yim, S.~Park, and J.~Kim, ``Joint fine-tuning in deep
  neural networks for facial expression recognition,'' in \emph{Proceedings of
  the IEEE International Conference on Computer Vision}, 2015, pp. 2983--2991.

\bibitem{khorrami2015deep}
P.~Khorrami, T.~Paine, and T.~Huang, ``Do deep neural networks learn facial
  action units when doing expression recognition?'' in \emph{Proceedings of the
  IEEE International Conference on Computer Vision Workshops}, 2015, pp.
  19--27.

\bibitem{liu2014deeply}
M.~Liu, S.~Li, S.~Shan, R.~Wang, and X.~Chen, ``Deeply learning deformable
  facial action parts model for dynamic expression analysis,'' in \emph{Asian
  Conference on Computer Vision}.\hskip 1em plus 0.5em minus 0.4em\relax
  Springer, 2014, pp. 143--157.

\bibitem{Goodfellow:NIPS2014}
I.~Goodfellow, J.~Pouget-Abadie, M.~Mirza, B.~Xu, D.~Warde-Farley, S.~Ozair,
  A.~Courville, and Y.~Bengio, ``Generative adversarial nets,'' in
  \emph{Advances in Neural Information Processing Systems 27}, 2014, pp.
  2672--2680.

\bibitem{zhaoMGL15}
J.~Zhao, M.~Mathieu, R.~Goroshin, and Y.~Lecun, ``{Stacked What-Where
  Auto-encoders},'' \emph{arXiv}, 2015.

\bibitem{wang2013capturing}
Z.~Wang, S.~Wang, and Q.~Ji, ``Capturing complex spatio-temporal relations
  among facial muscles for facial expression recognition,'' in
  \emph{Proceedings of the IEEE Conference on Computer Vision and Pattern
  Recognition}, 2013, pp. 3422--3429.

\bibitem{Bengio:2013}
Y.~Bengio, L.~Yao, G.~Alain, and P.~Vincent, ``Generalized denoising
  auto-encoders as generative models,'' in \emph{Proceedings of the 26th
  International Conference on Neural Information Processing Systems}, ser.
  NIPS'13, 2013, pp. 899--907.

\bibitem{hadsell2006dimensionality}
R.~Hadsell, S.~Chopra, and Y.~LeCun, ``Dimensionality reduction by learning an
  invariant mapping,'' in \emph{2006 IEEE Computer Society Conference on
  Computer Vision and Pattern Recognition (CVPR'06)}, vol.~2.\hskip 1em plus
  0.5em minus 0.4em\relax IEEE, 2006, pp. 1735--1742.

\bibitem{guo2012dynamic}
Y.~Guo, G.~Zhao, and M.~Pietik{\"a}inen, ``Dynamic facial expression
  recognition using longitudinal facial expression atlases,'' in \emph{Computer
  Vision--ECCV 2012}.\hskip 1em plus 0.5em minus 0.4em\relax Springer, 2012,
  pp. 631--644.

\bibitem{jain2011facial}
S.~Jain, C.~Hu, and J.~K. Aggarwal, ``Facial expression recognition with
  temporal modeling of shapes,'' in \emph{Computer Vision Workshops (ICCV
  Workshops), 2011 IEEE International Conference on}.\hskip 1em plus 0.5em
  minus 0.4em\relax IEEE, 2011, pp. 1642--1649.

\bibitem{conf/iccvw/PtuchaTS11}
R.~W. Ptucha, G.~Tsagkatakis, and A.~E. Savakis, ``Manifold based sparse
  representation for robust expression recognition without neutral
  subtraction.'' in \emph{ICCV Workshops}.\hskip 1em plus 0.5em minus
  0.4em\relax IEEE, 2011, pp. 2136--2143.

\bibitem{zhao2011oulucasia}
G.~Zhao, X.~Huang, M.~Taini, S.~Z. Li, and M.~Pietik{\"a}inen, ``Facial
  expression recognition from near-infrared videos,'' \emph{Image and Vision
  Computing}, vol.~29, no.~9, pp. 607--619, 2011.

\bibitem{zhong2012learning}
L.~Zhong, Q.~Liu, P.~Yang, B.~Liu, J.~Huang, and D.~N. Metaxas, ``Learning
  active facial patches for expression analysis,'' in \emph{Computer Vision and
  Pattern Recognition (CVPR), 2012 IEEE Conference on}.\hskip 1em plus 0.5em
  minus 0.4em\relax IEEE, 2012, pp. 2562--2569.

\bibitem{Li_2017_CVPR}
S.~Li, W.~Deng, and J.~Du, ``Reliable crowdsourcing and deep
  locality-preserving learning for expression recognition in the wild,'' in
  \emph{The IEEE Conference on Computer Vision and Pattern Recognition (CVPR)},
  July 2017.

\bibitem{GOODFELLOW201559}
I.~J. Goodfellow, D.~Erhan, P.~L. Carrier, A.~Courville, M.~Mirza, B.~Hamner,
  W.~Cukierski, Y.~Tang, D.~Thaler, D.-H. Lee, Y.~Zhou, C.~Ramaiah, F.~Feng,
  R.~Li, X.~Wang, D.~Athanasakis, J.~Shawe-Taylor, M.~Milakov, J.~Park,
  R.~Ionescu, M.~Popescu, C.~Grozea, J.~Bergstra, J.~Xie, L.~Romaszko, B.~Xu,
  Z.~Chuang, and Y.~Bengio, ``Challenges in representation learning: A report
  on three machine learning contests,'' \emph{Neural Networks}, vol.~64, pp. 59
  -- 63, 2015.

\bibitem{Emotiw2015}
A.~Dhall, O.~Ramana~Murthy, R.~Goecke, J.~Joshi, and T.~Gedeon, ``Video and
  image based emotion recognition challenges in the wild: Emotiw 2015,'' in
  \emph{Proceedings of the 2015 ACM on International Conference on Multimodal
  Interaction}, ser. ICMI '15.\hskip 1em plus 0.5em minus 0.4em\relax ACM,
  2015, pp. 423--426.

\bibitem{Yi2014CASIA}
S.~L. Dong~Yi, Zhen~Lei and S.~Z. Li, ``Learning face representation from
  scratch,'' in \emph{arXiv preprint arXiv:1411.7923. 2014}.

\bibitem{lucey2010CKextended}
P.~Lucey, J.~F. Cohn, T.~Kanade, J.~Saragih, Z.~Ambadar, and I.~Matthews, ``The
  extended cohn-kanade dataset (ck+): A complete dataset for action unit and
  emotion-specified expression,'' in \emph{2010 IEEE Computer Society
  Conference on Computer Vision and Pattern Recognition-Workshops}.\hskip 1em
  plus 0.5em minus 0.4em\relax IEEE, 2010, pp. 94--101.

\bibitem{valstar2010induced}
M.~Valstar and M.~Pantic, ``Induced disgust, happiness and surprise: an
  addition to the mmi facial expression database,'' in \emph{Proc. 3rd Intern.
  Workshop on EMOTION (satellite of LREC): Corpora for Research on Emotion and
  Affect}, 2010, p.~65.

\bibitem{noh2015learning}
H.~Noh, S.~Hong, and B.~Han, ``Learning deconvolution network for semantic
  segmentation,'' in \emph{Proceedings of the IEEE International Conference on
  Computer Vision}, 2015, pp. 1520--1528.

\bibitem{ding2017facenet2expnet}
H.~Ding, S.~K. Zhou, and R.~Chellappa, ``Facenet2expnet: Regularizing a deep
  face recognition net for expression recognition,'' in \emph{Automatic Face
  and Gesture Recognition (FG 2017), 2017 12th IEEE International Conference
  on}, 2017, pp. 118--126.

\bibitem{Klaser_aspatio-temporal2008}
A.~Klaser, M.~Marszałek, and C.~Schmid, ``A spatio-temporal descriptor based
  on 3d-gradients,'' in \emph{British Machine Vision Conference}, ser.
  BMVC’08, 2008.

\bibitem{maaten2008visualizing}
L.~v.~d. Maaten and G.~Hinton, ``Visualizing data using t-sne,'' \emph{Journal
  of Machine Learning Research}, vol.~9, no. Nov, pp. 2579--2605, 2008.

\bibitem{KingmaW13}
D.~P. Kingma and M.~Welling, ``Auto-encoding variational bayes,'' \emph{CoRR},
  vol. abs/1312.6114, 2013.

\bibitem{RadfordMC15}
A.~Radford, L.~Metz, and S.~Chintala, ``Unsupervised representation learning
  with deep convolutional generative adversarial networks,'' in
  \emph{International Conference on Learning Representations}, ser. ICLR2016,
  2016.

\bibitem{Nguyen_2015_CVPR}
A.~Nguyen, J.~Yosinski, and J.~Clune, ``Deep neural networks are easily fooled:
  High confidence predictions for unrecognizable images,'' in \emph{The IEEE
  Conference on Computer Vision and Pattern Recognition (CVPR)}, June 2015.

\bibitem{fer_wacv2016}
A.~Mollahosseini, D.~Chan, and M.~H. Mahoor, ``Going deeper in facial
  expression recognition using deep neural networks,'' in \emph{2016 IEEE
  Winter Conference on Applications of Computer Vision (WACV)}, March 2016, pp.
  1--10.

\bibitem{johnson84extensionslipschitz}
W.~Johnson and J.~Lindenstrauss, ``Extensions of {L}ipschitz mappings into a
  {H}ilbert space,'' in \emph{Conference in modern analysis and probability
  (New Haven, Conn., 1982)}, ser. Contemporary Mathematics.\hskip 1em plus
  0.5em minus 0.4em\relax American Mathematical Society, 1984, vol.~26, pp.
  189--206.

\end{thebibliography}
\end{document}